\documentclass[conference]{IEEEtran}
\pdfoutput=1
\usepackage{colortbl}
\usepackage{xspace}
\usepackage{todonotes}
\PassOptionsToPackage{hyphens}{url}\usepackage{hyperref}
\usepackage{csquotes}
\usepackage[skip=3pt]{caption}
\usepackage{mdwlist}
\usepackage{paralist}
\usepackage{tabularx}
\usepackage{subcaption}
\usepackage{graphicx}
\usepackage{amsmath,amssymb}
\usepackage{algorithm}
\usepackage{algpseudocode}
\usepackage{listings}
\usepackage{courier}
\usepackage{multirow}
\usepackage{booktabs}
\usepackage{balance} 
\usepackage{rotating}

\def\addauthnote#1#2{%
	\expandafter\def\csname#1\endcsname##1{\todo[inline,color=#2]{#1: ##1}\xspace}
	\expandafter\def\csname#1m\endcsname##1{\todo[color=#2]{#1: ##1}\xspace} }

\graphicspath{ {./} }

\newcommand\eg{\emph{e.g.}\xspace}
\newcommand\ie{\emph{i.e.}\xspace}
\newcommand\gym{\textsc{Gym FC}\xspace}

\newcommand\trials{2\xspace}
\newcommand\ci{95\%\xspace}

\newcommand\thresholdband{10\%\xspace}
\newcommand\thresholdriselow{10\%\xspace}
\newcommand\thresholdrisehigh{90\%\xspace}

\newcommand\ppoaa{0}
\newcommand\ppoab{0}
\newcommand\ppoac{0}
\newcommand\ppoad{0}
\newcommand\ppoae{0}
\newcommand\ppoaf{0}
\newcommand\ppoag{0}
\newcommand\ppoah{0}
\newcommand\ppoai{0}
\newcommand\ppoaj{0}
\newcommand\ppoak{0}
\newcommand\ppoal{0}

\newcommand\ppoba{0}
\newcommand\ppobb{0}
\newcommand\ppobc{0}
\newcommand\ppobd{0}
\newcommand\ppobe{0}
\newcommand\ppobf{0}
\newcommand\ppobg{0}
\newcommand\ppobh{0}
\newcommand\ppobi{0}
\newcommand\ppobj{0}
\newcommand\ppobk{0}
\newcommand\ppobl{0}

\newcommand\ppoca{0}
\newcommand\ppocb{0}
\newcommand\ppocc{0}
\newcommand\ppocd{0}
\newcommand\ppoce{0}
\newcommand\ppocf{0}
\newcommand\ppocg{0}
\newcommand\ppoch{0}
\newcommand\ppoci{0}
\newcommand\ppocj{0}
\newcommand\ppock{0}
\newcommand\ppocl{0}

\newcommand\trpoaa{0}
\newcommand\trpoab{0}
\newcommand\trpoac{0}
\newcommand\trpoad{0}
\newcommand\trpoae{0}
\newcommand\trpoaf{0}
\newcommand\trpoag{0}
\newcommand\trpoah{0}
\newcommand\trpoai{0}
\newcommand\trpoaj{0}
\newcommand\trpoak{0}
\newcommand\trpoal{0}

\newcommand\trpoba{0}
\newcommand\trpobb{0}
\newcommand\trpobc{0}
\newcommand\trpobd{0}
\newcommand\trpobe{0}
\newcommand\trpobf{0}
\newcommand\trpobg{0}
\newcommand\trpobh{0}
\newcommand\trpobi{0}
\newcommand\trpobj{0}
\newcommand\trpobk{0}
\newcommand\trpobl{0}

\newcommand\trpoca{0}
\newcommand\trpocb{0}
\newcommand\trpocc{0}
\newcommand\trpocd{0}
\newcommand\trpoce{0}
\newcommand\trpocf{0}
\newcommand\trpocg{0}
\newcommand\trpoch{0}
\newcommand\trpoci{0}
\newcommand\trpocj{0}
\newcommand\trpock{0}
\newcommand\trpocl{0}

\newcommand\ddpgaa{0}
\newcommand\ddpgab{0}
\newcommand\ddpgac{0}
\newcommand\ddpgad{0}
\newcommand\ddpgae{0}
\newcommand\ddpgaf{0}
\newcommand\ddpgag{0}
\newcommand\ddpgah{0}
\newcommand\ddpgai{0}
\newcommand\ddpgaj{0}
\newcommand\ddpgak{0}
\newcommand\ddpgal{0}

\newcommand\ddpgba{0}
\newcommand\ddpgbb{0}
\newcommand\ddpgbc{0}
\newcommand\ddpgbd{0}
\newcommand\ddpgbe{0}
\newcommand\ddpgbf{0}
\newcommand\ddpgbg{0}
\newcommand\ddpgbh{0}
\newcommand\ddpgbi{0}
\newcommand\ddpgbj{0}
\newcommand\ddpgbk{0}
\newcommand\ddpgbl{0}

\newcommand\ddpgca{0}
\newcommand\ddpgcb{0}
\newcommand\ddpgcc{0}
\newcommand\ddpgcd{0}
\newcommand\ddpgce{0}
\newcommand\ddpgcf{0}
\newcommand\ddpgcg{0}
\newcommand\ddpgch{0}
\newcommand\ddpgci{0}
\newcommand\ddpgcj{0}
\newcommand\ddpgck{0}
\newcommand\ddpgcl{0}

\newcommand\pida{0}
\newcommand\pidb{0}
\newcommand\pidc{0}
\newcommand\pidd{0}
\newcommand\pide{0}
\newcommand\pidf{0}
\newcommand\pidg{0}
\newcommand\pidh{0}
\newcommand\pidi{0}
\newcommand\pidj{0}
\newcommand\pidk{0}
\newcommand\pidl{0}

\newcommand\rtddpg{33 hours\xspace}

\newcommand\rtppo{9 hours\xspace}

\newcommand\rttrpo{13 hours\xspace}

\newcommand\extra{(Appendix~\ref{sec:continuous})\xspace}

\begin{document}
\title{Reinforcement Learning for UAV Attitude Control} 

\author{
		\IEEEauthorblockN{William Koch, Renato Mancuso, Richard West, Azer 
			Bestavros}
			\IEEEauthorblockA{Boston University\\
				Boston, MA 02215
					\\\{wfkoch, rmancuso, richwest, best\}@bu.edu}
					}

\maketitle
\thispagestyle{plain}
\pagestyle{plain}
\begin{abstract}

Autopilot systems are typically composed of an ``inner loop''
providing stability and control, while an ``outer loop'' is
responsible for mission-level objectives, e.g. way-point navigation.
Autopilot systems for UAVs are predominately implemented using
Proportional, Integral Derivative~(PID) control systems, which have
demonstrated exceptional performance in stable environments. However
more sophisticated control is required to operate in unpredictable,
and harsh environments.  Intelligent flight control systems is an
active area of research addressing limitations of PID control most recently 
through the use of reinforcement learning~(RL) which has had success in other 
applications such as robotics.  However
previous work has focused primarily on using RL at the mission-level
controller.
In this work, we investigate the performance and accuracy of the inner
control loop providing attitude control when using intelligent flight
control systems trained with the state-of-the-art RL algorithms, Deep
Deterministic Gradient Policy~(DDGP), Trust Region Policy
Optimization~(TRPO) and Proximal Policy Optimization~(PPO).  To
investigate these unknowns we first developed an open-source
high-fidelity simulation environment to train a flight controller
attitude control of a quadrotor through RL. We then use our
environment to compare their performance to that of a PID controller
to identify if using RL is appropriate in high-precision,
time-critical flight control.
\end{abstract}

\section{Introduction}

Over the last decade there has been an uptrend in the popularity of
Unmanned Aerial Vehicles~(UAVs). In particular, quadrotors have
received significant attention in the research community where a
significant number of seminal results and applications has been proposed and experimented.
This recent growth is primarily attributed to the drop in cost of
onboard sensors, actuators and small-scale embedded computing
platforms. Despite the significant progress, flight control is still
considered an open research topic. On the one hand, flight control
inherently implies the ability to perform highly time-sensitive
sensory data acquisition, processing and computation of forces to
apply to the aircraft actuators. On the other hand, it is desirable
that UAV flight controllers are able to tolerate faults; adapt to
changes in the payload and/or the environment; and to optimize flight
trajectory, to name a few.

Autopilot systems for UAVs are typically composed of an ``inner loop" 
responsible for aircraft stabilization and control, and an ``outer loop" to 
provide mission level objectives (\eg way-point navigation).  Flight control 
systems for UAVs are predominately implemented using the Proportional, Integral 
Derivative~(PID) control systems. PIDs have demonstrated exceptional 
performance in many circumstances, including in the context of drone racing, 
where precision and agility are key. In stable 
environments a PID controller exhibits close-to-ideal performance. 
When exposed to unknown dynamics (\eg wind, variable payloads, voltage 
sag, etc), however, a PID controller can be far from optimal~\cite{maleki2016reliable}.
For next-generation flight control systems to be intelligent, a way needs to be devised to
incorporate adaptability to mutable dynamics and environment.

The development of intelligent flight control systems is an active area of 
research~\cite{santoso2017state}, specifically through the use of artificial 
neural networks which are an attractive option given they are universal 
approximators and resistant to noise~\cite{miglino1995evolving}. 

Online learning methods~(\eg \cite{dierks2010output}) have the advantage of 
learning the aircraft dynamics in real-time. The main limitation with online learning 
is that the flight control system is 
only knowledgeable of its past experiences. It follows that its performances are limited when 
exposed to a new event.  Training models offline using {\bf supervised learning} is 
problematic as data is expensive to obtain and derived from inaccurate 
representations of the underlying aircraft dynamics (\eg flight data from a 
similar aircraft using PID control) which can lead to suboptimal control policies~\cite{bobtsov2016hybrid,shepherd2010robust,williams2005flight}.
To construct high-performance intelligent flight control systems it is necessary 
to use a hybrid approach. First, accurate offline models are used to construct a
baseline controller, while online learning provides fine tuning and real-time adaptation. 

An alternative to supervised learning for creating offline models is known as 
{\bf reinforcement learning (RL)}.  In RL an agent is given a reward for every action  
it makes in an environment with the objective to maximize the rewards over time.  
Using RL it is possible to develop optimal control policies for a UAV without 
making any assumptions about the aircraft dynamics. 
Recent work has shown RL to be effective for UAV autopilots, providing adequate 
path tracking~\cite{hwangbo2017control}. Nonetheless, previous  work on intelligent 
flight control systems has primarily focused on guidance and navigation. 
It remains unclear what level of control accuracy can be achieved 
when using intelligent control for time-sensitive attitude control --- \ie the 
``inner loop".  Determining the achievable level of accuracy 
is critical in establishing for what applications intelligent flight 
control is indeed suitable. For instance, high 
precision and accuracy is necessary for proximity or indoor flight. But accuracy may be 
sacrificed in larger outdoor spaces where adaptability is of the utmost importance 
due to the unpredictability of the environment.

In this paper we study the accuracy and precision of attitude control provided 
by intelligent flight controllers trained using RL. While we specifically focus on the 
creation of controllers for the Iris quadcopter~\cite{iris}, the 
methods developed hereby apply to a wide range of multi-rotor UAVs, and can also be extended 
to fixed-wing aircraft. 
We develop a novel training environment called \gym with the use of a high 
fidelity physics simulator for the agent to learn attitude control.  \gym is an 
OpenAI Environment~\cite{brockman2016openai} providing a common interface for 
researchers to develop intelligent flight control systems.  The simulated 
environment consists of an Iris quadcopter digital replica or \textit{digital 
twin}~\cite{gabor2016simulation} with the intention of eventually be used to 
transfer the trained controller to physical hardware.  Controllers are trained 
using state-of-the-art RL algorithms: Deep Deterministic Gradient Policy~(DDGP), 
Trust Region
Policy Optimization~(TRPO), and Proximal Policy Optimization~(PPO). We then 
compare the performance of our synthesized controllers with that of a PID 
controller.  Our evaluation finds that controllers trained using PPO outperform 
PID control and are capable of exceptional performance. 
To summarize, this paper makes the following contributions: 

\begin{itemize}
\item \gym, an open source~\cite{gymfc} environment for developing intelligent 
	attitude flight controller providing the research community a tool to
	progress performance.

\item A learning architecture for attitude control utilizing digital twinning 
	concepts for minimal effort when transferring trained controllers into 
	hardware.

\item An evaluation for state-of-the-art RL algorithms, such as Deep Deterministic 
      Gradient Policy~(DDGP), Trust Region Policy Optimization~(TRPO),
	  and Proximal Policy Optimization~(PPO), learning policies for
      aircraft attitude control. As a first work in this direction,
      our evaluation also establishes a baseline for future work.

\item An analysis of intelligent flight control performance developed with RL 
      compared to traditional PID control.

\end{itemize}

The remainder of this paper is organized as follows. In Section~\ref{sec:bg} we 
provide an overview of the quadcopter flight dynamics and reinforcement 
learning. Next, in Section~\ref{sec:related} we briefly survey existing literature on intelligent 
flight control. In Section~\ref{sec:env} we present our training environment and 
use this environment to evaluate RL performance for flight control in 
Section~\ref{sec:eval}. Finally Section~\ref{sec:conclusion} concludes the paper
and provides a number of future research directions.

\section{Background}
\label{sec:bg}
In this section we provide a brief overview of quadcopter flight 
dynamics required to understand this work, and an introduction to developing flight control 
systems with reinforcement learning.   

\subsection{Quadcopter Flight Dynamics}
\label{sec:flightdyn}
A quadcopter is an aircraft with six degrees of freedom (DOF), three
rotational and three translational. With four control inputs (one to
each motor) this results in an under-actuated system that requires an
onboard computer to compute motor signals to provide stable flight. We
indicate with $\omega_i, i \in {1, \dots, M}$ the rotation speed of
each rotor where $M=4$ is the total number of motors for a quadcopter. These 
have a direct impact on the resulting Euler angles
$\phi, \theta, \psi$, \ie roll, pitch, yaw respectively which provide rotation 
in $D=3$ dimensions.  Moreover,
they produce a certain amount of upward thrust, indicated with $f$.

\begin{figure}
	\centering

	\begin{subfigure}[b]{0.5\textwidth}
	{\includegraphics[trim=35 405 20 100, clip, width=\textwidth]{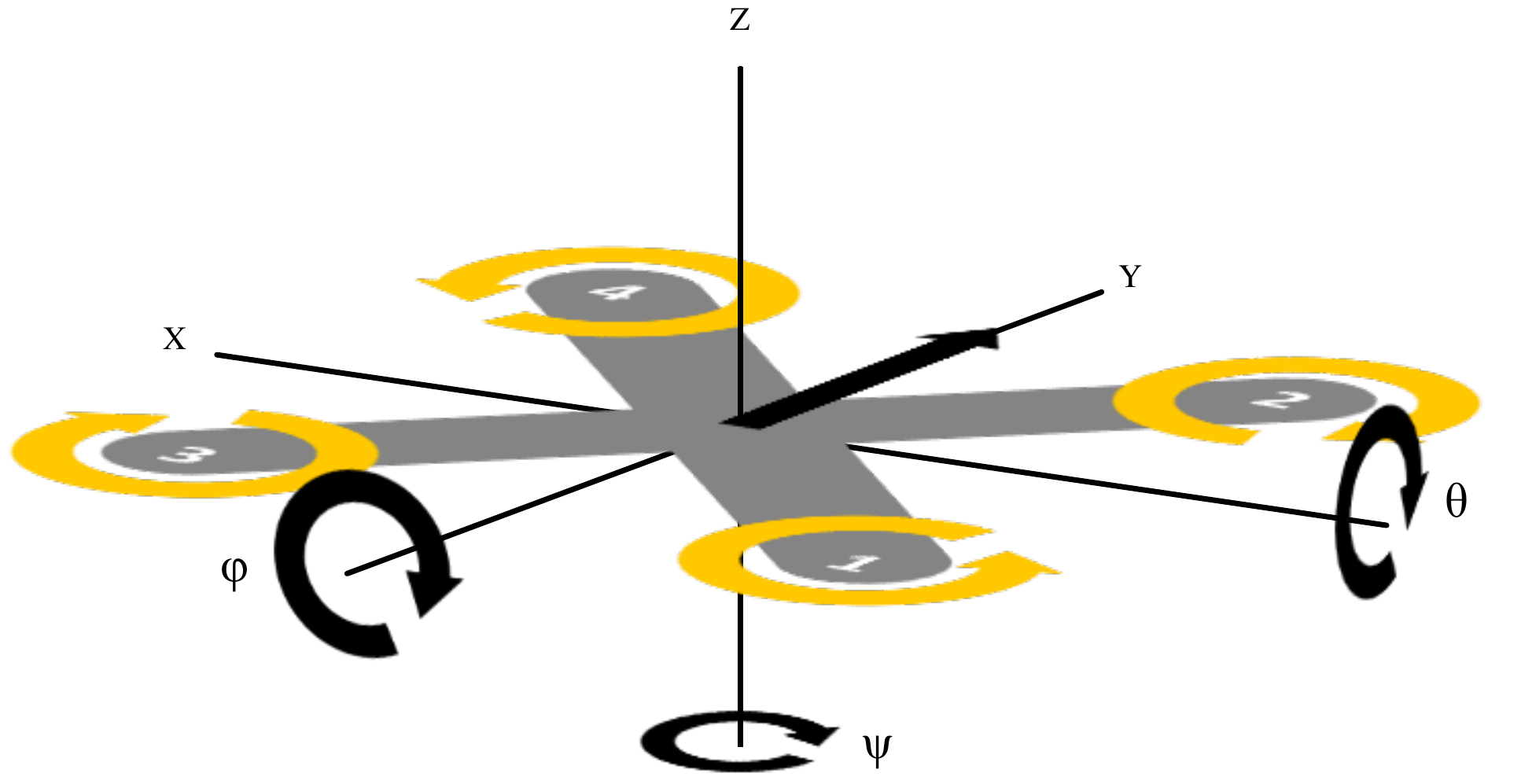}}
	\caption{Axis of rotation}
	\label{fig:axis}
	\end{subfigure} \\	
	
	\begin{subfigure}[b]{0.5\textwidth}
		\centering
		\begin{subfigure}[b]{0.3\textwidth}
		{\includegraphics[trim=30 660 470 18, clip, 
		width=\textwidth]{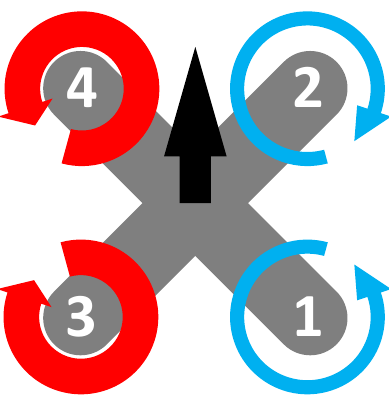}}
		\caption{Roll right}
		\label{fig:fd:roll}
		\end{subfigure}
		\begin{subfigure}[b]{0.3\textwidth}
		{\includegraphics[trim=28 660 473 20, clip,
		width=\textwidth]{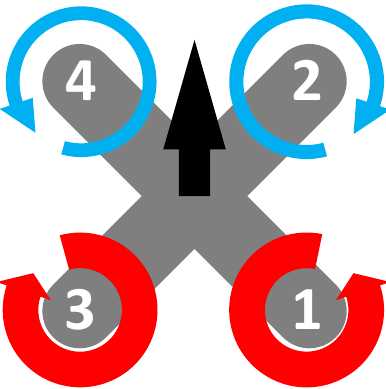}}
		\caption{Pitch forward}
		\label{fig:fd:pitch}
		\end{subfigure}
		\begin{subfigure}[b]{0.3\textwidth}
		{\includegraphics[trim=30 660 470 20, clip, width=\textwidth]{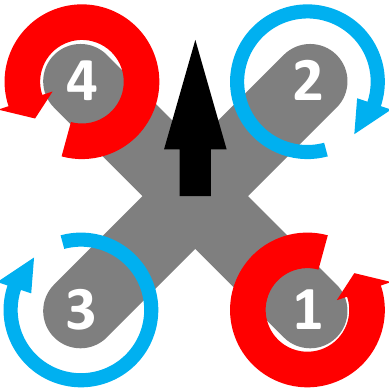}}
		\caption{Yaw  clockwise}
		\label{fig:fd:yaw}
		\end{subfigure}

	\end{subfigure}
\label{fig:quadcopter}
\end{figure}

The aerodynamic effect that each $\omega_i$ produces depends upon the
configuration of the motors.  The most popular configuration is
an ``X'' configuration, depicted in Figure~\ref{fig:axis} which has the motors 
mounted in an ``X" formation relative to what is considered the front of the 
aircraft.  This
configuration provides more stability compared to a ``+'' configuration which in 
contrast has its motor configuration rotated an additional $45^{\circ}$ along 
the z-axis. This is due to the differences in torque generated along each axis 
of rotation in respect to the distance of the motor from the axis.  The aerodynamic affect $u$ that 
each rotor speed $\omega_i$ has on
thrust and Euler angles, is given by:
\begin{align}
u_f &= b ( \omega_1^2 + \omega_2^2 + \omega_3^2 + \omega_4^2)\\
u_\phi &= b ( \omega_1^2 + \omega_2^2 - \omega_3^2 - \omega_4^2)\\
u_\theta &= b ( \omega_1^2 - \omega_2^2 + \omega_3^2 - \omega_4^2)\\
u_\psi &= b ( \omega_1^2 - \omega_2^2 - \omega_3^2 + \omega_4^2)
\label{eq:qc_dynamics}
\end{align}
where $u_f, u_\phi, u_\theta, u_\psi$ is the thrust, roll, pitch, and
yaw effect respectively, while $b$ is a thrust factor that captures
propeller geometry and frame characteristics.  For further details
about the mathematical models of quadcopter dynamics please refer
to~\cite{bouabdallah2004design}.

To perform a rotational movement the velocity of each rotor is manipulated 
according to the relationship expressed in Equation~\ref{eq:qc_dynamics} and as illustrated in 
Figure~\ref{fig:fd:roll}, \ref{fig:fd:pitch}, \ref{fig:fd:yaw}.  For example, to 
roll right (Figure~\ref{fig:fd:roll}) more thrust is delivered to motors~3 and~4. 
Yaw (Figure~\ref{fig:fd:yaw}) is not achieved directly through difference in 
thrust generated by the rotor as roll and pitch are, but instead through a
difference in torque in the rotation speed of rotors spinning in opposite directions.  
For example, as shown in Figure~\ref{fig:fd:yaw}, higher rotation speed for rotors~1 and~4 
allow the aircraft to yaw clockwise. Because a net positive torque counter-clockwise causes the 
aircraft to rotate clockwise due to Newton's second law of motion. 

Attitude, in respect to orientation of a quadcopter, can be expressed by its  
angular velocities of each axis $\Omega=[\Omega_\phi, \Omega_\theta, 
\Omega_\psi]$. The objective of attitude control is to compute the required 
motor signals to achieve some desired attitude $\Omega^*$.

In autopilot systems attitude control is executed as an inner control loop  and 
is time-sensitive.  Once the desired attitude is achieved, translational 
movement (in the X, Y, Z direction) is accomplished by applying thrust 
proportional to each motor.

In commercial quadcopter, the vast if not all use PID attitude control. A PID 
controller is a linear feedback controller expressed mathematically as,
\begin{equation}
	u(t) = K_p e(t) + K_i \int_0^t e(\tau) d \tau + K_d \frac{d e(t)}{dt}
\end{equation}
where $K_p, K_i, K_d$ are configurable constant gains and $u(t)$ is the control 
signal. The effect of each term can be thought of as the P term considers the 
current error, the I term considers the history of errors and the D term 
estimates the future error. For attitude control in a quadcopter aircraft there 
is PID control for each roll, pitch and yaw axis. At each cycle in the inner 
loop, each PID sum is computed for each axis and then these values are 
translated into the amount of power to deliver to each motor through a process 
called \textit{mixing}. Mixing uses a table consisting of constants describing 
the geometry of the frame to determine how the axis control signals are summed 
based on the torques that will be generated by the length of each arm~(recall 
differences between ``X" and ``+" frames).  The control signal for each motor  
$y_i$ is loosely defined as,
\begin{equation}
y_i=f\left( m_{(i,\phi)} u_\phi + m_{(i,\theta)} u_\theta +m_{(i,\psi)} u_\psi 
\right)
\label{eq:mix}
\end{equation} where $m_{(i,\phi)},m_{(i,\theta)},m_{(i,\psi)}$ are the mixer 
values for motor $i$ and $f$ is the throttle coefficient.

\subsection{Reinforcement Learning}
\label{sec:rl}

In this work we consider a neural network flight controller as an agent  
interacting with an Iris quadcopter~\cite{iris} in a high fidelity physics 
simulated  environment $\mathcal{E}$, more specifically using the
Gazebo simulator~\cite{koenig2004design}.

At each discrete time-step $t$, the agent receives an observation $x_t$ from the 
environments  consisting of the angular velocity error of each axis $e = 
\Omega^* - \Omega$ and the angular velocity of each rotor $\omega_i$ which are 
obtained from the quadcopter's gyroscope and electronic speed controller (ESC) 
respectively. These observations are in the continuous observation spaces $x_t 
\in \mathbb{R}^{(M+D)}$. 
Once the observation  is received, the agent executes an action $a_t$ within 
$\mathcal{E}$.  In return the agent receives a single numerical reward $r_t$ 
indicating the performance of this action.  The action is also in a
continuous action space $a_t \in  \mathbb{R}^M$ and corresponds to the four 
control signals sent to each motor.  
Because the agent is only receiving this sensor data it is unaware of the 
physical environment and the aircraft dynamics and therefore $\mathcal{E}$ is 
only partially observed by the agent.  Motivated by \cite{mnih2013playing} we 
consider the state to be a sequence of the past observations and actions $s_t = 
x_i, a_i, \dots, a_{t-1}, x_t$.

The interaction between the agent and $\mathcal{E}$ is formally defined as a 
Markov decision processes~(MDP) where the state transitions are defined as the 
probability of transitioning to state $s'$ given the current state and action 
are $s$ and $a$ respectively, $Pr\{s_{t+1} = s' | s_t = s, a_t = a\}$.  The 
behavior of the agent is defined by its policy $\pi$ which is essentially a 
mapping of what action should be taken for a particular state.  The objective of 
the agent is to maximize the returned reward overtime to develop an optimal 
policy.  We welcome the reader to refer to \cite{sutton1998reinforcement} for 
further details on reinforcement learning.

Up until recently control in continuous action space was considered  difficult 
for RL. Significant progress has been made combining the power of neural 
networks with RL. In this work we elected to use  Deep Deterministic Gradient 
Policy~(DDGP) \cite{lillicrap2015continuous} and Trust Region Policy 
Optimization~(TRPO)~\cite{schulman2015trust} due to the recent use of these 
algorithms for quadcopter navigation control~\cite{hwangbo2017control}.  DDPG 
provides improvement to Deep Q-Network~(DQN)~\cite{mnih2013playing}
for the continuous action domain. It employs an actor-critic architecture using 
two neural networks for each actor and critic.  It is also model-free algorithm 
meaning it can learn the policy without having to first generate a model. TRPO 
is similar to natural gradient policy methods however this method guarantees 
monotonic improvements. We additionally include a third algorithm for our 
analysis called  Proximal Policy Optimization~(PPO) \cite{schulman2017proximal}.  
PPO  is known to out perform other state-of-the-art  methods in challenging 
environments. PPO is also a policy gradient method and has similarities to TRPO 
while being easier to implement and tune.

\section{Related Work}
\label{sec:related}
\label{sec:related}
Aviation has a rich history in flight control dating back to the 1960s. During 
this time supersonic aircraft were being developed which demanded more 
sophisticated dynamic flight control than what a static linear controller could 
provide.  Gain scheduling~\cite{leith2000survey} was developed allowing multiple 
linear controllers of different configurations to be used in designated 
operating regions. This however was inflexible and insufficient for handling the 
nonlinear dynamics at high speeds but paved way for adaptive control. For a 
period of time many experimental adaptive controllers were being tested but were 
unstable.  Later advances were made to increase stability with model reference 
adaptive control (MRAC)~\cite{whitaker1958design}, and $L_1$ 
\cite{hovakimyan2011mathcal} which provided reference models during adaptation.  
As the cost of small scale embedded computing platforms dropped, intelligent 
flight control options became realistic and have been actively researched over 
the last decade to design flight control solutions that are able to 
\textit{adapt}, but also to \textit{learn}.

As performance demands for UAVs continues to increase we are beginning to see 
signs that flight control history repeats itself. The popular high performance 
drone racing firmware Betaflight~\cite{betaflight} has recently added gain 
scheduler to adjust PID gains depending on throttle and voltage levels. 
Intelligent PID flight control~\cite{fatan2013adaptive} methods have been 
proposed in which PID gains are dynamically updated online providing adaptive 
control as the environment changes. However these solutions still  
inherit disadvantages associated with PID control, such as integral windup,  
need for mixing, and most significantly, they are feedback controllers and 
therefore inherently \textit{reactive}.  On the other hand feedforward control (or 
predictive control) is \textit{proactive}, and allows the controller to output 
control signals before an error occur. For feedforward control, a model of the system must exist.  
Learning-based intelligent control has been proposed to develop models of the 
aircraft for predictive control using artificial neural networks.  

Notable work by Dierks et. al.~\cite{dierks2010output} proposes an intelligent 
flight control system constructed with neural networks to learn the quadcopter 
dynamics, online, to navigate along a specified path. This method allows the 
aircraft to adapt in real-time to external disturbances and unmodeled dynamics.  
Matlab simulations demonstrate that their approach outperforms a PID controller in the 
presence of unknown dynamics, specifically in regards to control effort 
required to track the desired trajectory. Nonetheless the proposed approach does 
requires prior knowledge of the aircraft mass and moments of inertia to estimate 
velocities.  Online learning is an essential component to constructing a 
complete intelligent flight control system. It is fundamental however to develop 
accurate offline models to account for uncertainties encountered during online 
learning~\cite{santoso2017state}.  
To build offline models, previous work has used supervised learning to train 
intelligent flight control systems using a variety of data sources such as test 
trajectories~\cite{bobtsov2016hybrid}, and PID step responses 
\cite{shepherd2010robust}.
The limitation of this approach is that training data may not accurately reflect 
the underlying dynamics. In general, supervised learning on its own is not ideal 
for interactive problems such as control~\cite{sutton1998reinforcement}.

Reinforcement learning has similar goals to adaptive control in which a policy 
improves overtime interacting with its environment. The first use of 
reinforcement learning in  quadcopter control was presented by Waslander et.  
al.~\cite{waslander2005multi} for altitude control. The authors developed a 
model-based reinforcement learning algorithm to search for an optimal control 
policy. The controller was rewarded for accurate tracking and damping. Their 
design provided significant improvements in stabilization in comparison to linear 
control system.
More recently Hwangbo et. al.~\cite{hwangbo2017control} has used reinforcement 
learning for quadcopter control, particularly for navigation control. They 
developed a novel deterministic on-policy learning algorithm that outperformed 
TRPO~\cite{schulman2015trust} and DDPG~\cite{lillicrap2015continuous} in regards 
to training time. Furthermore the authors validated their results in the real 
world, transferring their simulated model to a physical quadcopter.  Path 
tracking turned out to be adequate.  Notably, the authors discovered major 
differences transferring from simulation to the real world.  Known as the 
\textit{reality gap}, transferring from simulation to the real-world has been 
researched extensively as being problematic without taking additional steps to 
increase realism in the simulator~\cite{jakobi1995noise,miglino1995evolving}.

The vast majority of prior work has focused on performance of navigation and 
guidance. There is limited and insufficient data justifying the accuracy and 
precision of neural-network-based intelligent attitude flight control and none 
to our knowledge for controllers trained using RL. Furthermore this work uses 
physics simulations in contrast to mathematical models of the aircraft and 
environments used in aforementioned prior work for increased realism. {\bf The 
goal of this work is to provide a platform for training attitude controllers 
with RL, and to provide performance baselines in regards to attitude controller 
accuracy.}

\section{Environment}
\label{sec:env}
In this section we describe our learning environment \gym for developing 
intelligent flight control systems using RL.  The goal of proposed environment is to
allow the agent to learn attitude control of an aircraft with only the 
knowledge of the number of actuators.  \gym includes both an {\bf episodic task} and a 
{\bf continuous task}. In an episodic task, the agent is required to learn a policy for 
responding to individual angular velocity commands. This allows the agents to 
learn the step response from rest for a given command, allowing its performance 
to be accurately measured. Episodic tasks however are not reflective of realistic
flight conditions. For this reason, in a continuous task, pulses with random widths 
and amplitudes are continuously generated, and correspond to angular velocity set-points. 
The agent must respond accordingly and track the desired target over time.
In Section~\ref{sec:eval} we evaluate our synthesized controllers via episodic tasks, 
but we have strong experimental evidence that training via episodic tasks produces
controllers that behave correctly in continuous tasks as well~\extra.

\begin{figure}
\centering
{\includegraphics[trim=0 573 273 0, clip, width=\columnwidth]{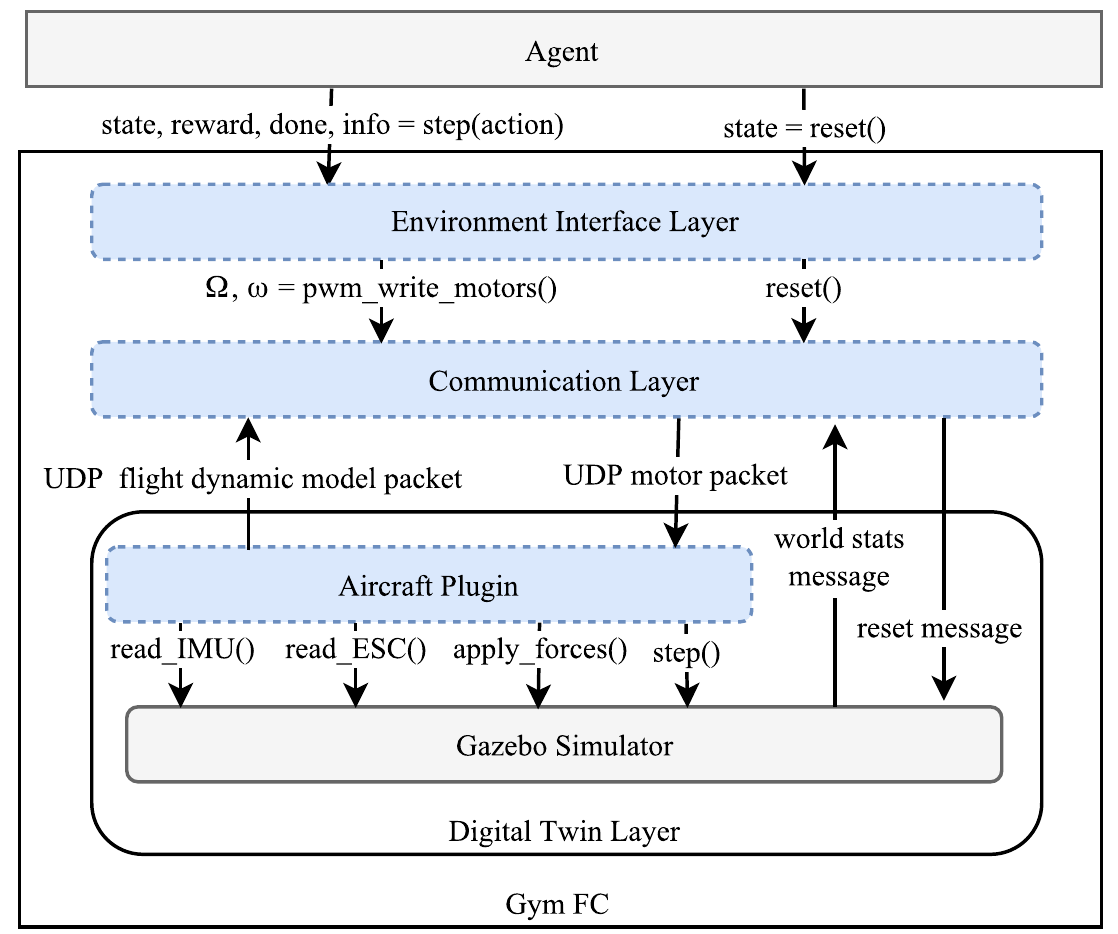}}
\caption{Overview of environment architecture, \gym. Blue blocks with dashed 
borders are implementations developed for this work. }
\label{fig:env}
\end{figure}

\gym has a multi-layer hierarchical architecture composed of three layers: (i) a 
digital twin layer, (ii) a communication layer, and (iii) an agent-environment interface layer.  
This design decision was made to clearly establish roles and allow layer 
implementations to change (\eg to use a different simulator) without affecting 
other layers as long as the layer-to-layer interfaces remain intact.  A high 
level overview of the environment architecture is illustrated in 
Figure~\ref{fig:env}.  We will now discuss in greater detail each layer with a 
bottom-up approach.

\subsection{Digital Twin Layer}
At the heart of the learning environment is a high fidelity physics simulator 
which provides functionality and realism that is hard to achieve with an
abstract mathematical model of the aircraft and environment. One of the primary
design goals of \gym is to minimize the effort required to transfer a controller
from the learning environment into the final platform.
For this reason, the simulated environment exposes identical interfaces to  
actuators and sensors as they would exist in the physical world.  In the ideal 
case the agent should not be able to distinguish between interaction with the 
simulated world (\ie its digital twin) and its hardware counter part.  In this 
work we use the Gazebo simulator~\cite{koenig2004design} in light of its 
maturity, flexibility, extensive documentation, and active community.  

In a nutshell, the {\bf digital twin layer} is defined by (i) the
simulated world, and (ii) its interfaces to the above communication
layer (see Figure~\ref{fig:env}).

\begin{figure}
	\centering
	{\includegraphics[trim=45 0 65 30, clip, width=0.48\textwidth]{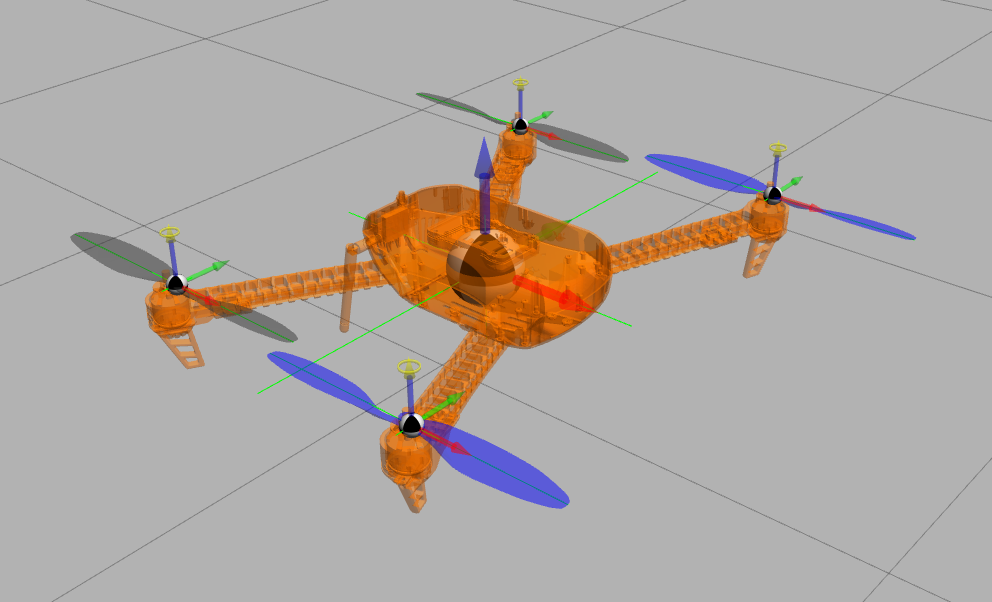}}
	\caption{The Iris quadcopter in Gazebo one meter above the ground.  The body 
	is transparent to show where the center of mass is linked as a ball joint to 
	the world.  Arrows represent the various joints used in the model. }
	\label{fig:iris}
\end{figure}

\textbf{Simulated World} The simulated world is constructed specifically for UAV 
attitude control in mind.  The technique we developed allows attitude control to 
be accomplished independently of guidance and/or navigation control.  This is 
achieved by fixing the center of mass of the aircraft to a ball joint in the 
world, allowing it to rotate freely in any direction, which would be impractical 
if not impossible to achieved in the real world due to gimbal lock and friction 
of such an apparatus. In this work the aircraft to be controlled in the 
environment is modeled off of the Iris quadcopter~\cite{iris}  with a weight of 
1.5~Kg, and 550~mm motor-to-motor distance. An illustration of the quadcopter in 
the environment is displayed in Figure~\ref{fig:iris}. Note during training 
Gazebo runs in headless mode without this user interface to increase simulation 
speed.  This architecture however can be used with any multicopter as long as a 
digital twin can be constructed.  Helicopters and multicopters represent 
excellent candidates for our setup because they can achieve a full range of 
rotations along all the three axes.  This is typically not the case with 
fixed-wing aircraft.  Our design can however be expanded to support fixed-wing 
by simulating airflow over the control surfaces for attitude control.  Gazebo 
already integrates a set of tools to perform airflow simulation.

\textbf{Interface} The digital twin layer provides two command interfaces to the 
communication layer:  simulation reset and motor update. Simulation reset 
commands are supported by Gazebo's API and are not part of our implementation.  
Motor updates are provided by a UDP server. We hereby discuss our approach to 
developing this interface.

In order to keep synchronicity between the simulated world and the controller
of the digital twin, the pace at which simulation should progress is directly 
enforced. This is possible by controlling the simulator step-by-step. In our 
initial approach, Gazebo's Google Protobuf~\cite{protobuf} API was used, with a 
specific message to progress by a single simulation step.  By subscribing to 
status messages (which include the current simulation step) it is possible to 
determine when a step has completed and to ensure synchronization.  However as 
we attempted to increase the rate of advertising step messages, we discovered 
that the rate of status messages is capped at 5~Hz.
Such a limitation introduces a consistent bottleneck in the simulation/learning pipeline. 
Furthermore it was found that Gazebo silently drops messages it cannot process.

A set of important modifications were made to increase experiment
throughput.  The key idea was to allow motor update commands to
directly drive the simulation clock.  By default Gazebo comes
pre-installed with an ArduPilot Arducopter~\cite{ardupilot} plugin to
receive motor updates through a UDP server. These motor updates are in
the form of a pulse width modulation (PWM) signals. At the same time,
sensor readings from the inertial measurement unit (IMU) on board the
aircraft is sent over a second UDP channel.  Arducopter is an open
source multicopter firmware and its plugin was developed to support
software in the loop (SITL).

We derived our Aircraft Plugin from the Arducopter plugin with the
following modifications (as well as those discussed in
Section~\ref{sec:comm}). Upon receiving a motor command, the motor
forces are updated as normal but then a simulation step is executed.
Sensor data is read and then sent back as a response to the client
over the same UDP channel.  In addition to the IMU sensor data we also
simulate sensor data obtained from the electronic speed
controller~(ESC). The ESC provides the angular velocities of each
rotor, which are relayed to the client too.  {\bf Implementing our Aircraft 
Plugin with this approach successfully allowed us to work around
the limitations of the Google Protobuf API and increased step throughput
by over 200 times.}

\subsection{Communication Layer}
\label{sec:comm}
The communication layer is positioned in between the digital twin and the 
agent-environment interface.  This layer manages the low-level communication channel to 
the aircraft and simulation control. The primary function of this layer is to 
export a high-level synchronized API to the higher layers for interacting with 
the digital twin which uses asynchronous communication protocols. This layer 
provides the commands \texttt{pwm\_write} and \texttt{reset} to the 
agent-environment interface layer.  

The function call \texttt{pwm\_write} takes as input a vector  of PWM values for 
each actuator, corresponding to the control input $u(t)$. These PWM values 
correspond to the same values that would be sent to an ESC on a physical UAV.    
The PWM values are translated to a normalized format expected by the Aircraft 
Plugin, and then packed into a UDP packet for transmission to the Aircraft Plugin UDP 
server.  The communication layer blocks until a response is received from the 
Aircraft Plugin, forcing synchronized writes for the above layers.  The UDP 
reply is unpacked and returned in response.  

During the learning process the simulated environment must be reset at the 
beginning of each learning episode. Ideally one could use the \texttt{gz} 
command line utility included with the Gazebo installation which is lightweight 
and does not require additional dependencies. Unfortunately there is a known 
socket handle leak~\cite{gzbug} that causes Gazebo to crash if the command is 
issued more than the maximum number of open files allowed by the operating 
system.  Given we are running thousands episodes during training this was not an 
option for us.  Instead we opted to use the Google Protobuffer interface so we 
did not have to deploy a patched version of the utility on our test servers. 
Because resets only occur at the beginning of a training session and are not in 
the critical processing loop using Google Protobuffers here is acceptable.

Upon start of the communication layer a connection is established with the 
Google Protobuff API server and we subscribe to world statistics messages which 
includes the current simulation iteration. To reset the simulator a world 
control message is adversed instructing the simulator to reset the simulation 
time. The communication layer blocks until it receives a world statistics 
message indicating the simulator has been reset and then returns back control to 
the agent-environment interface layer.  Note the world control message is only 
resetting the simulation time, not the entire simulator (\ie models and 
sensors). This is because we found that in some cases when a world control 
message was issued to perform a full reset the sensor data took a few additional 
iterations for reset. To ensure proper reset to the above layers this time reset 
message acts as a signalling mechanism to the Aircraft Plugin.  When the plugin 
detects a time reset has occurred it resets the whole simulator and most 
importantly steps the simulator until the sensor values have also reset ensuring 
above layers that when a new training session starts, reading sensor values 
accurately reflect the current state and not the previous state from stale 
values.

\subsection{Environment Interface Layer}

The topmost layer interfacing with the agent is the environment interface layer 
which  implements the OpenAI Gym~\cite{brockman2016openai} environment API.  
Each OpenAI Gym environment defines an observation space and an action space.  
These are used to inform the agent of the bounds to expect for environment 
observations and what are legal bounds for the action input, respectively.
As previously mentioned in Section~\ref{sec:rl} \gym is in both the continuous 
observation space and action space domain. The state is of size $m\times(M+D)$ 
where $m$ is the memory size indicating the number of past observations; $M = 4$ 
as we consider a four-motors configuration; and $D = 3$ since each measurement 
is taken in the 3 dimensions.
Each observation value is in $[-\infty : \infty]$. The action space 
is of size $M$ equivalent to the number of control actuators of the aircraft  (\ie 
four for a quadcopter), where each value is normalized between $[-1:1]$ to be 
compatible with most agents who squash their output using the $\tanh$ function.  

\gym implements two primary OpenAI functions, namely \texttt{reset} and 
\texttt{step}.  The \texttt{reset} function is called at the start of an episode 
to reset the environment and returns the initial environment state. This is 
also when the desired target angular velocity $\Omega^*$ or setpoint is 
computed.
The setpoint is randomly sampled from a uniform distribution between 
$[\Omega_{min}, \Omega_{max}]$. For the continuous task this is also set at a 
random interval of time. Selection of these bounds may refer to the desired 
operating region of the aircraft.  Although it is highly unlikely during normal 
operation that a quadcopter will be expected to reach the majority of these target 
angular velocities, the intention of these tasks are to push and stress the 
performance of the aircraft. 

The \texttt{step} function executes a single simulation step with the specified 
actions and returns to the agent the new state vector, together with a reward 
indicating how well the given action was performed.
Reward engineering can be challenging. If careful design is not performed, the derived 
policy may not reflect what was originally intended. Recall from 
Section~\ref{sec:rl} that the reward is ultimately what shapes the policy. For 
this work, with the goal of establishing a baseline of accuracy, we develop a 
reward to reflect the current angular velocity error (\ie $e = \Omega^* - 
\Omega$).  In the future \gym will be expanded to include additional 
environments aiding in the development of more complex policies particularity to 
showcase the advantages of using RL to adapt and learn.
We translate the current error $e_t$ at time $t$ into into a derived reward  
$r_t$ normalized between $[-1, 0]$ as follows,
\begin{equation}
r_t = -clip\left( sum(|\Omega^*_t - \Omega_t| )/ 3 \Omega_{max} \right)
\label{eq:sumabserr}
\end{equation}
where the $sum$  function sums the absolute value of the error of each axis, and 
the $clip$  function clips the result between the $[0,1]$ in cases where there 
is an overflow in the error. Since the reward is negative, it signifies a 
penalty, the agent maximizes the rewards (and thus minimizing error) overtime in 
order to track the target as accurately as possible. Rewards are normalized to 
provide standardization and stabilization during training \cite{normalize}. 

Additionally we also experimented with a variety of other rewards. We found 
sparse binary rewards\footnote{A reward structured so that $r_t=0$ if 
$sum(|e_t|) < threshold$, otherwise $r_t = -1$.} 
to give poor performance. We believe this to be due to complexity of 
quadcopter control. In the early stages of learning the agent explores its 
environment. However the event of randomly reaching the target angular velocity 
within some threshold was rare and thus did not provide the agent with enough 
information to converge. Conversely, we found that signalling at each timestep was 
best.  We also tried using the Euclidean norm of the error, quadratic error and 
other scalar values, all of which did not provide performance that could come close 
to what achieved with the sum of absolute errors (Equation~\ref{eq:sumabserr}).

\section{Evaluation}
\label{sec:eval}
In this section we present our evaluation on the accuracy of studied
neural-network-based attitude flight controllers trained with RL. Due
space limitations, we present evaluation and results only for episodic
tasks, as they are directly comparable to our baseline
(PID). Nonetheless, we have obtained strong experimental evidence that
agents trained using episodic tasks perform well in continuous
tasks~\extra. To our knowledge, this is the first RL baseline
conducted for quadcopter attitude control.

\subsection{Setup}

We evaluate the RL algorithms DDGP, TRPO, and PPO using the implementations in the
OpenAI Baselines project~\cite{baselines}. The goal of the OpenAI Baselines project 
is to establish a reference implementation of RL algorithms, providing baselines for 
researchers to compare approaches and build upon. Every algorithm is run with 
defaults except for the number of simulations steps which we increased to 10~million. 

The episodic task parameters were configured to run each episode for a maximum 
of 1~second of simulated time allowing enough time for the controller to 
respond to the command as well as additional time
to identify if a steady state has been reached.  The bounds the target angular 
 velocity is sampled from is set to $\Omega_{min}=-5.24$ rad/s, 
 $\Omega_{max}=5.24$ rad/s ($\pm$ 300 deg/s). These limits were constructed by 
 examining PID's performance to make sure we expressed physically feasible 
 constraints.  The max step size of the Gazebo simulator, which specifies the 
 duration of each physics update step was set to 1~ms to develop highly accurate 
 simulations. In other words, our physical world ``evolved'' at 1~kHz.
Training and evaluations were run on Ubuntu~16.04 with an eight-core 
i7-7700 CPU and an NVIDIA GeForce GT~730 graphics card.

For our PID controller, we ported the mixing and SITL implementation from 
Betaflight~\cite{betaflight} to Python to be compatible with \gym. 
The PID controller was first tuned using the classical Ziegler-Nichols 
method~\cite{ziegler1942optimum} and then manually adjusted to improve 
performance of the step response sampled around the midpoint $\pm\Omega_{max}/2$.
We obtained the following gains for each axis of rotation: 
$K_\phi = [2, 10, 0.005], K_\theta = [10, 10, 0.005], K_\psi = [4, 50, 0.0]$, where each vector 
contains to the $[K_p,K_i,K_d]$ (proportional, integrative, derivative) gains, respectively.
Next we measured the distances between the arms of the quadcopter to calculate 
the mixer values for each motor $m_i, i \in \{1, \ldots 4\}$.  Each vector $m_i$ 
is of the form $m_i = [ m_{(i,\phi)}, m_{(i,\theta)}, m_{(i,\psi)}]$, i.e.   
roll, pitch, and yaw (see Section~\ref{sec:flightdyn}). The final values were: 
$m_1 = [ -1.0,  0.598, -1.0 ]$, $m_2 = [ -0.927, -0.598,  1.0 ]$,  $m_3 = [ 1.0,  
0.598,  1.0 ]$ and lastly $m_4 = [ 0.927, -0.598, -1.0 ]$. The mix values and 
PID sums are then used to compute each motor signal $y_i$ according to 
Equation~\ref{eq:mix}, where $f=1$ for no additional throttle.  

To evaluate and compare the accuracy of the different algorithms we used a 
set of metrics. First, we define ``initial error'' as the distance between the rest velocities
and the current setpoint. A notion of progress toward the setpoint from rest can then be expressed as
the percentage of the initial error that has been ``corrected''. Correcting 0\% of the initial
error means that no progress has been made; while 100\% indicates that the setpoint
has been reached. Each metric value is independently computed for each axis. We hereby list our metrics.
\textbf{Success} captures the
number of experiments (in percentage) in which the controller eventually settles in 
an band within 90\% an 110\% of the initial error, \ie $\pm10\%$ from the setpoint. 
\textbf{Failure} captures the average percent error relative to the initial error 
after $t=500~ms$, for those experiments that do not make it in the $\pm10\%$ error band. 
The latter metric quantifies the magnitude of unacceptable controller performance.
The delay in the measurement ($t>500~ms$) is to exclude the rise regime. The underlying
assumption is that a steady state is reached before $500~ms$. 
\textbf{Rise} is the average time 
in milliseconds it takes the controller to go from \thresholdriselow to 
\thresholdrisehigh of the initial error. \textbf{Peak} is the max achieved angular 
velocity represented as a percentage relative to the initial error.  
Values greater than 100\% indicate overshoot, while values less than 100\% represent 
undershoot.
\textbf{Error} is the average sum of the absolute value error of each episode in 
$rad/s$. This provides a generic metric for performance. Our last 
metric is \textbf{Stability}, which captures how stable the response is halfway 
through the simulation, i.e. at $t>500ms$. Stability is calculated by taking the 
linear regression of the angular velocities and reporting the slope of the 
calculated line. Systems that are unstable have a non-zero slope.

\begin{figure*}[htp]
	\centering
	\begin{subfigure}[b]{0.3\textwidth}
		{\includegraphics[width=1.75in]{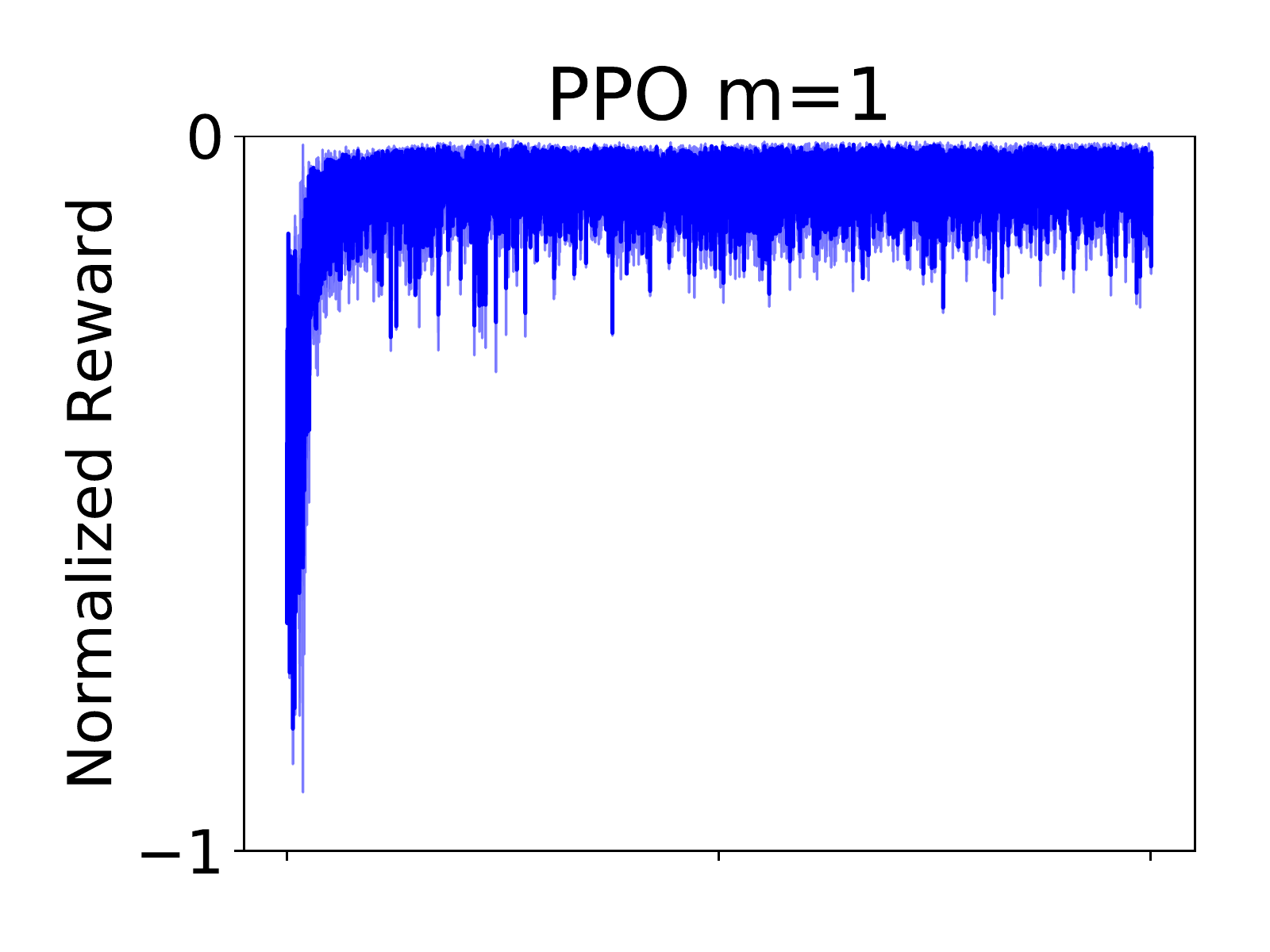}}
	\end{subfigure}
	\hfill
	\begin{subfigure}[b]{0.3\textwidth}
		{\includegraphics[width=1.75in]{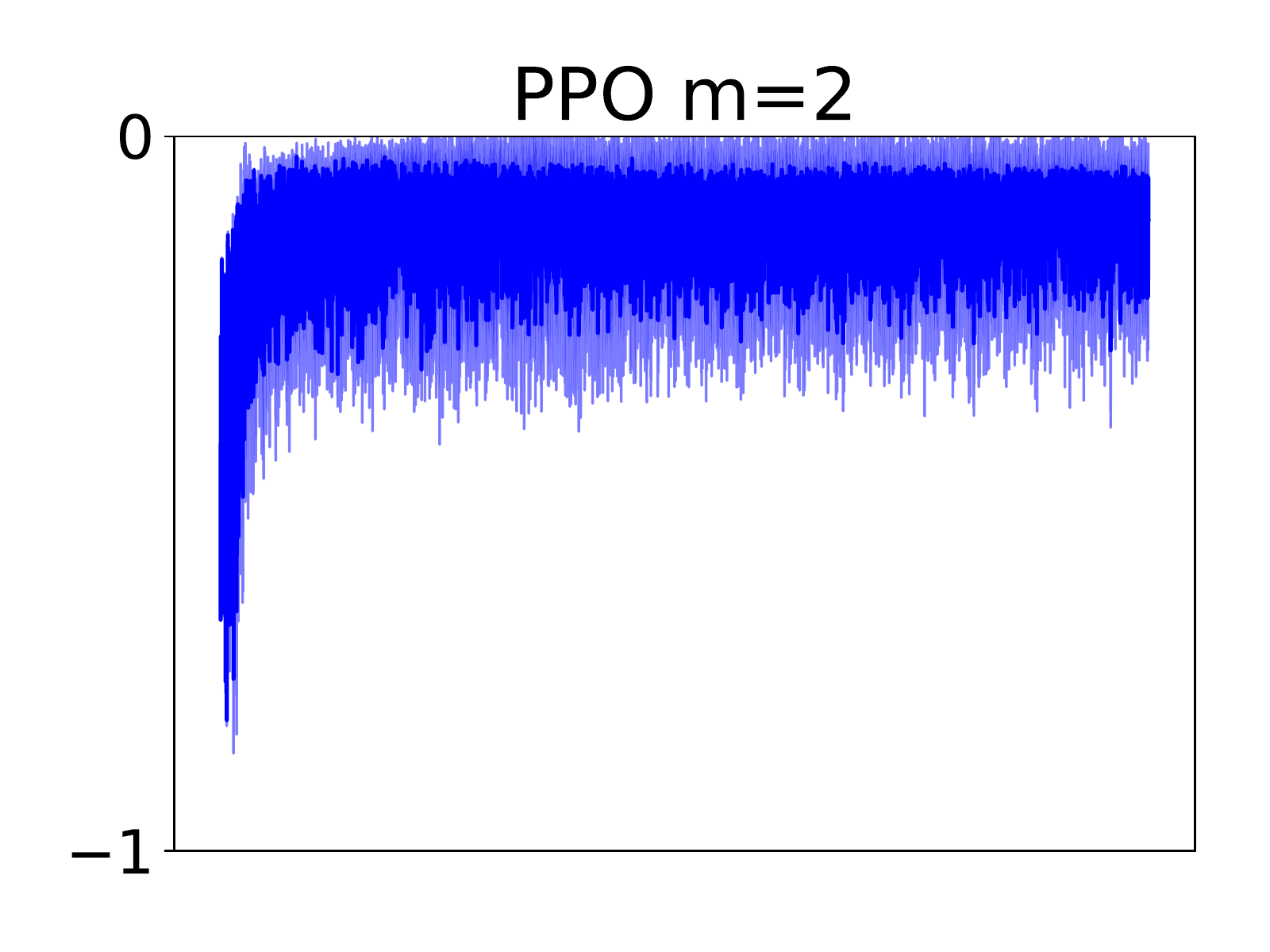}}
	\end{subfigure}
	\hfill
	\begin{subfigure}[b]{0.3\textwidth}
		{\includegraphics[width=1.75in]{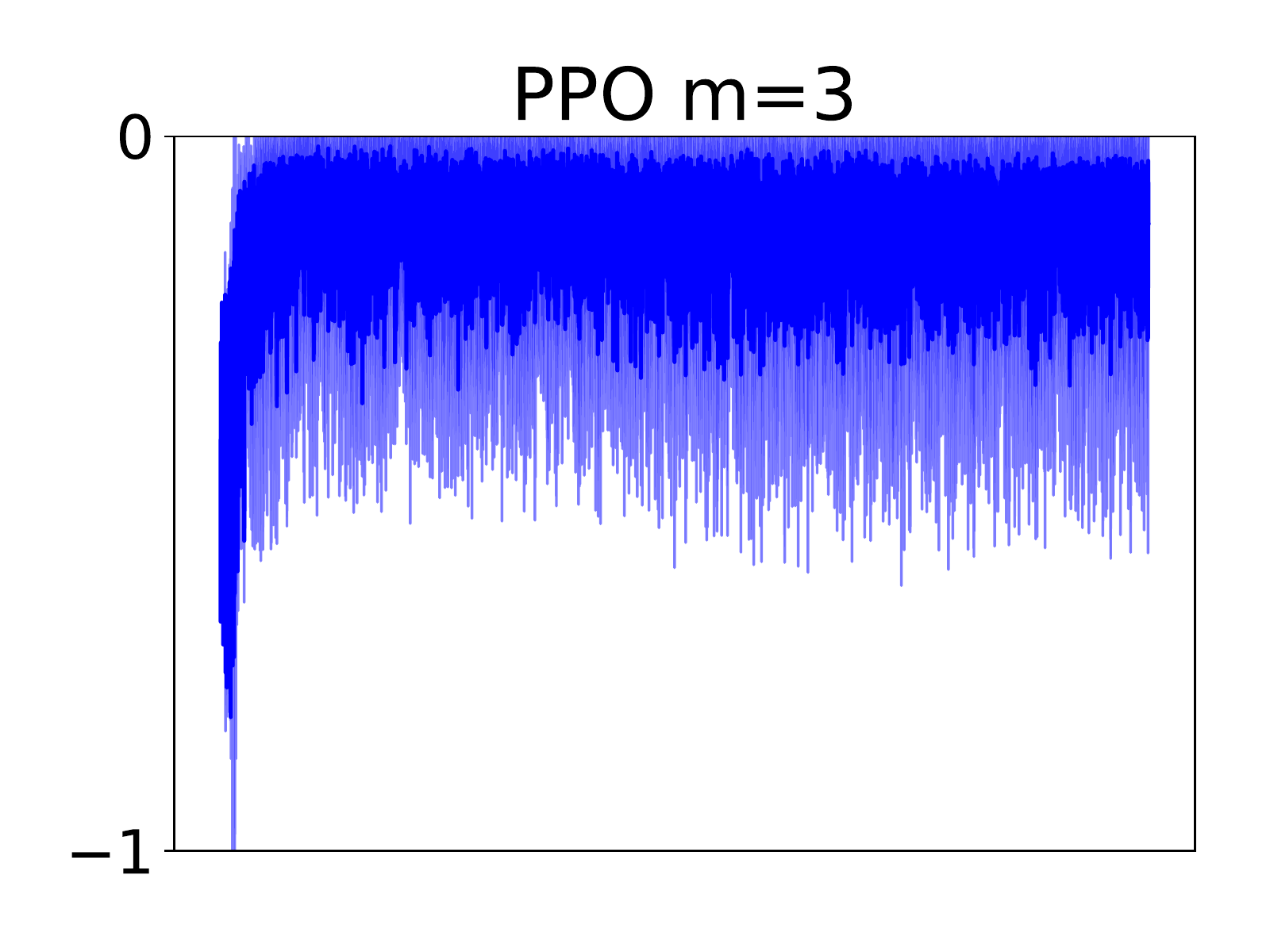}}
	\end{subfigure}
	\\[-2.0ex]
	\begin{subfigure}[b]{0.3\textwidth}
		{\includegraphics[width=1.75in]{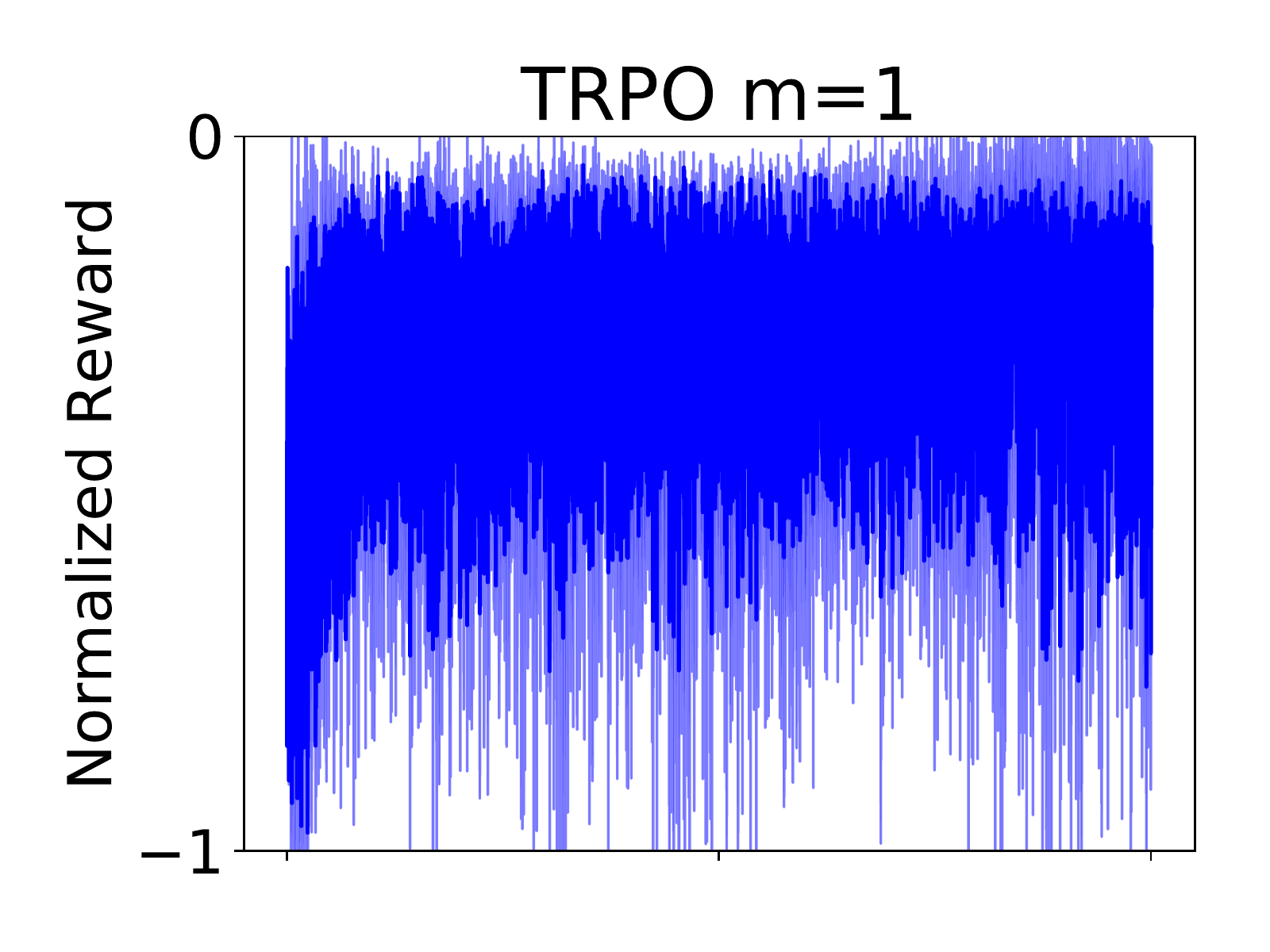}}
	\end{subfigure}
	\hfill
	\begin{subfigure}[b]{0.3\textwidth}
		{\includegraphics[width=1.75in]{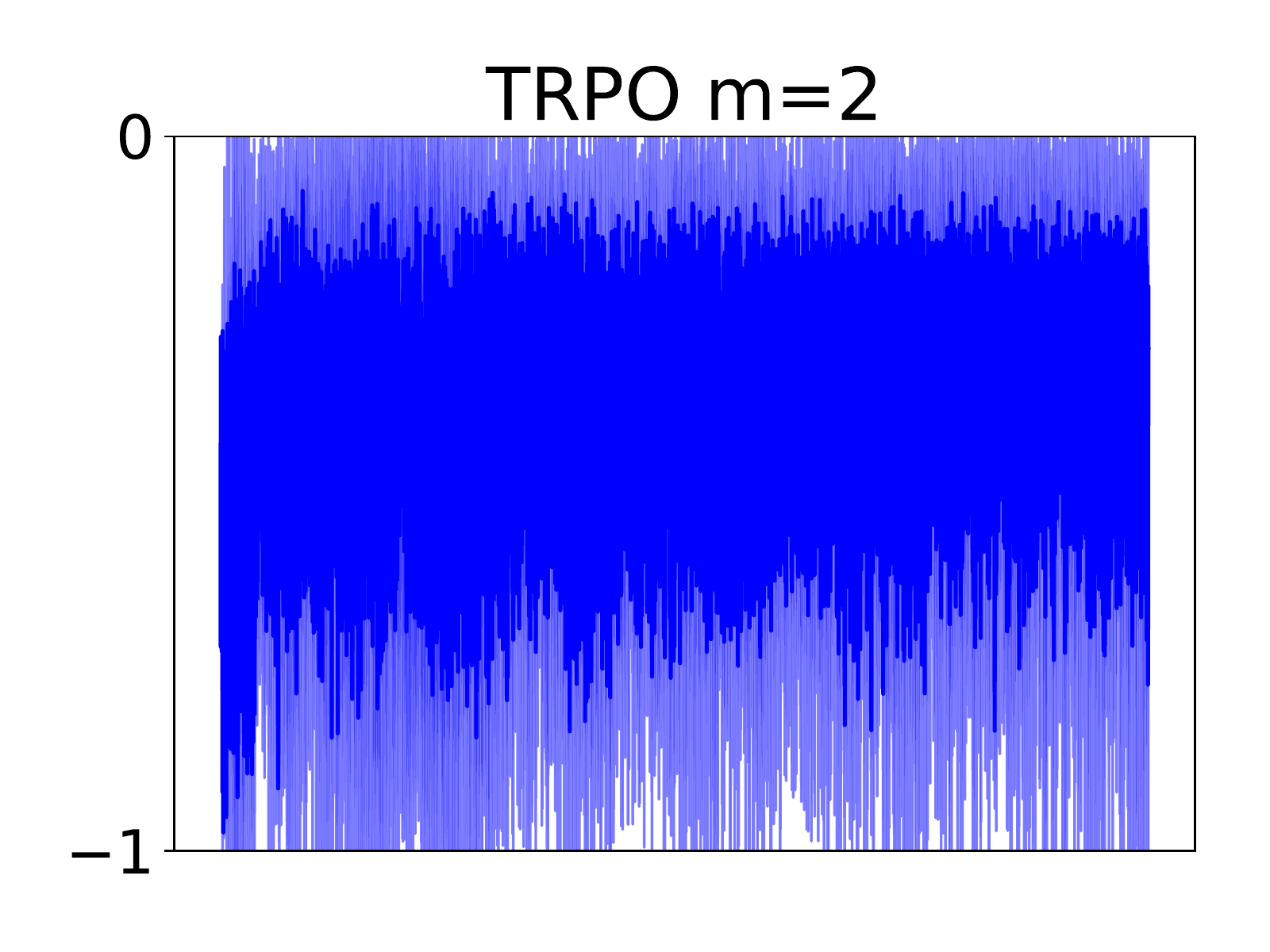}}
	\end{subfigure}
	\hfill
	\begin{subfigure}[b]{0.3\textwidth}
		{\includegraphics[width=1.75in]{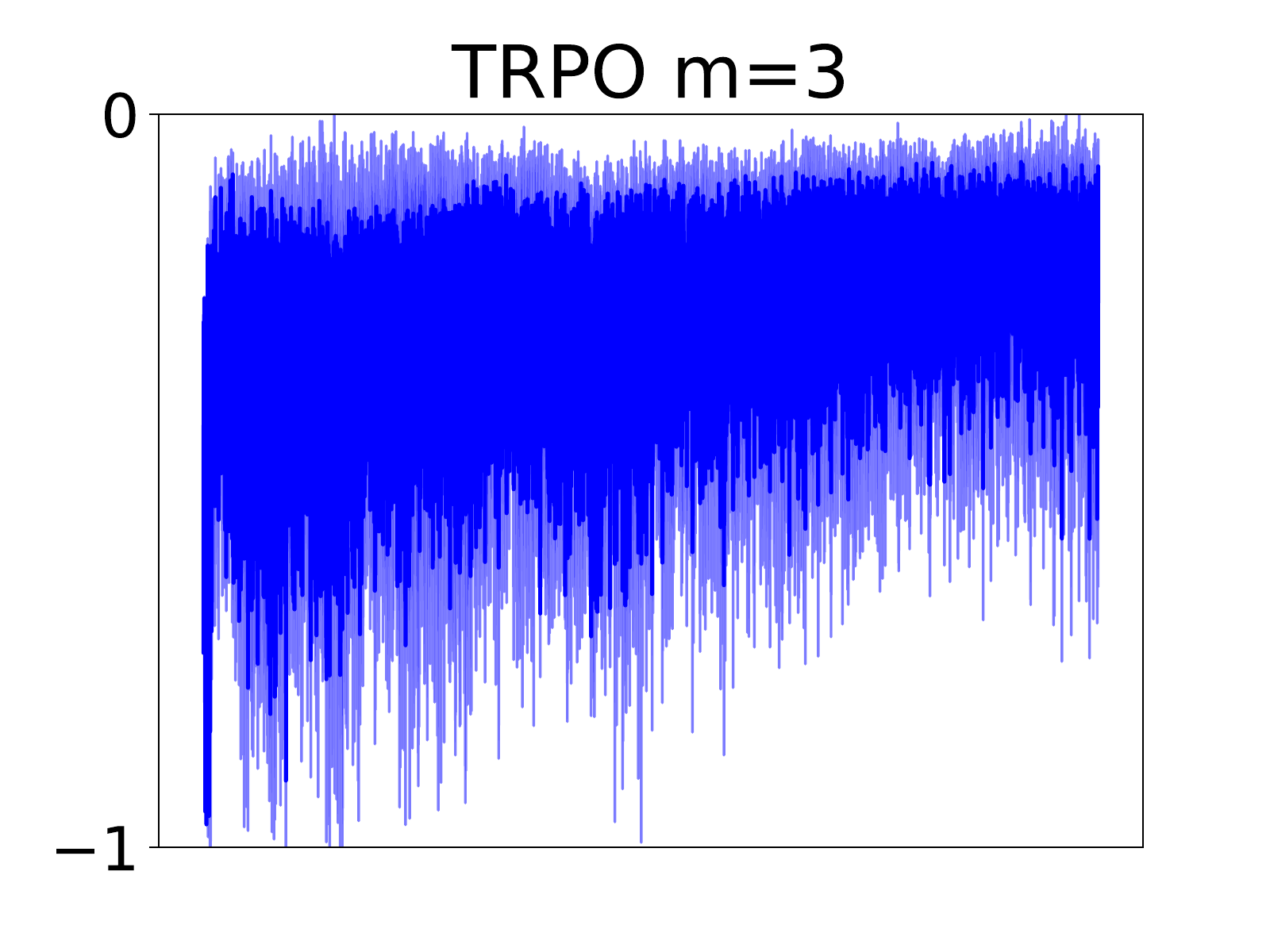}}
	\end{subfigure}
	\\[-2.0ex]
	\begin{subfigure}[b]{0.3\textwidth}
		{\includegraphics[width=1.75in]{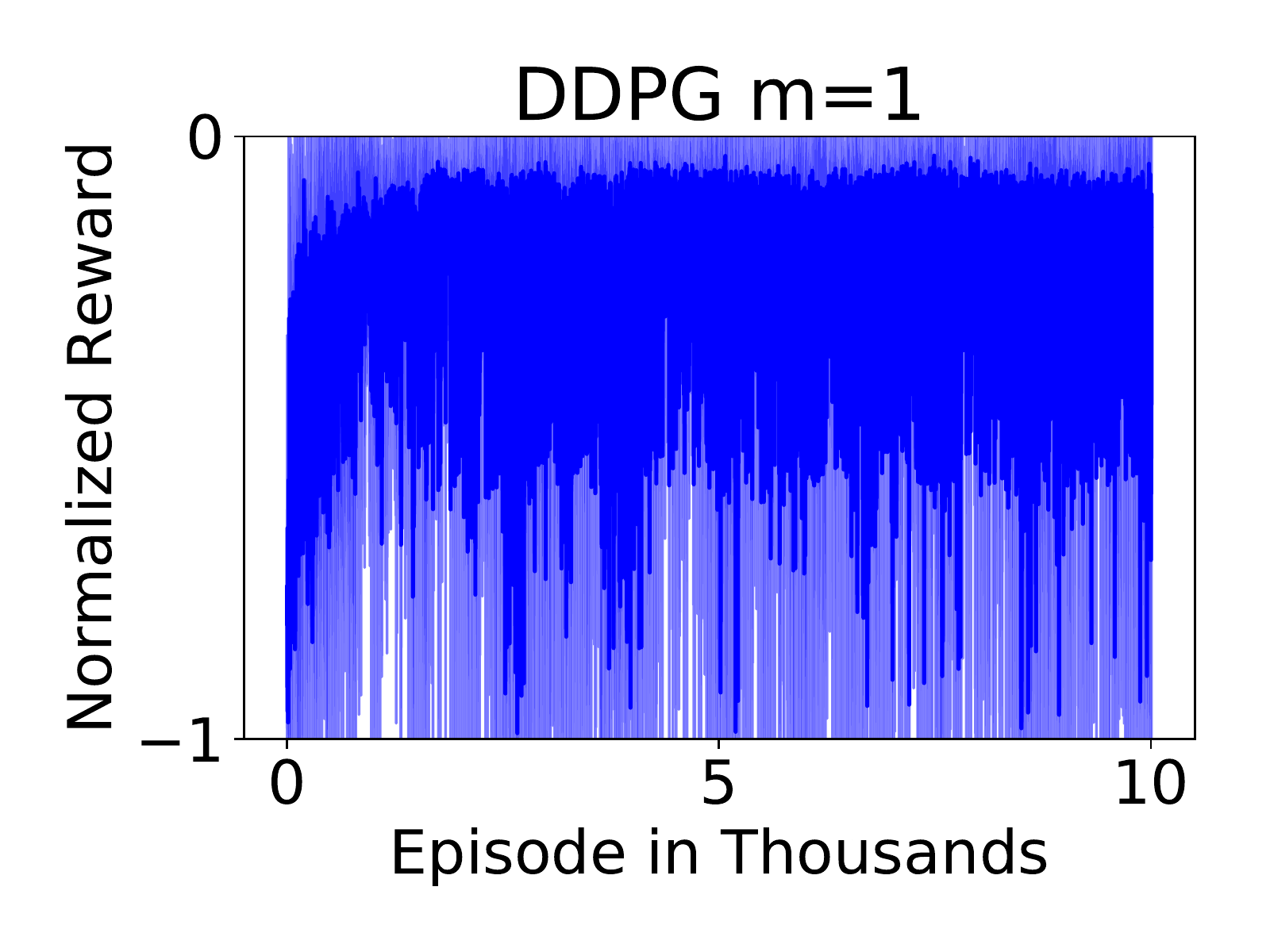}}
	\end{subfigure}
	\hfill
	\begin{subfigure}[b]{0.3\textwidth}
		{\includegraphics[width=1.75in]{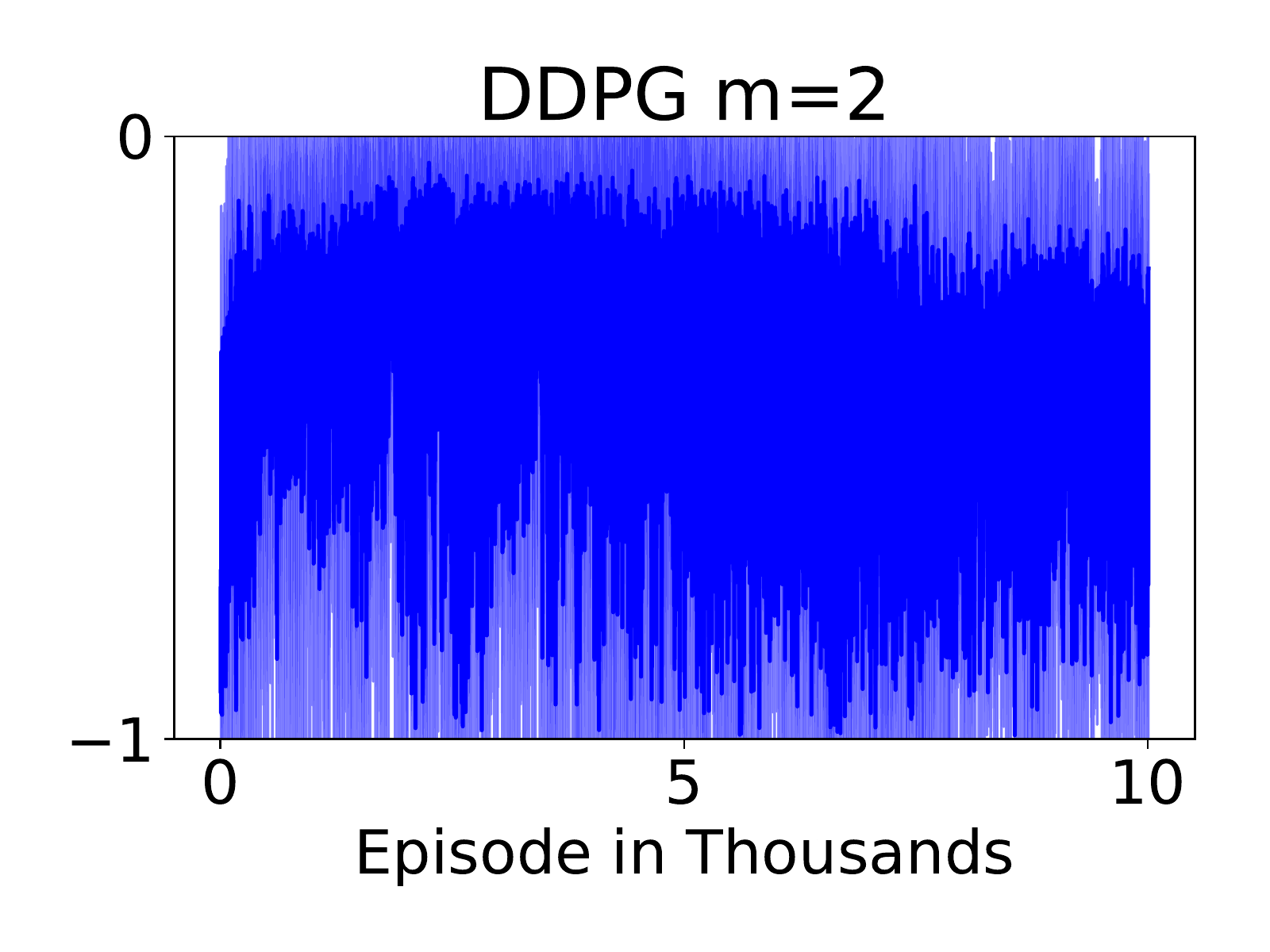}}
	\end{subfigure}
	\hfill
	\begin{subfigure}[b]{0.3\textwidth}
		{\includegraphics[width=1.75in]{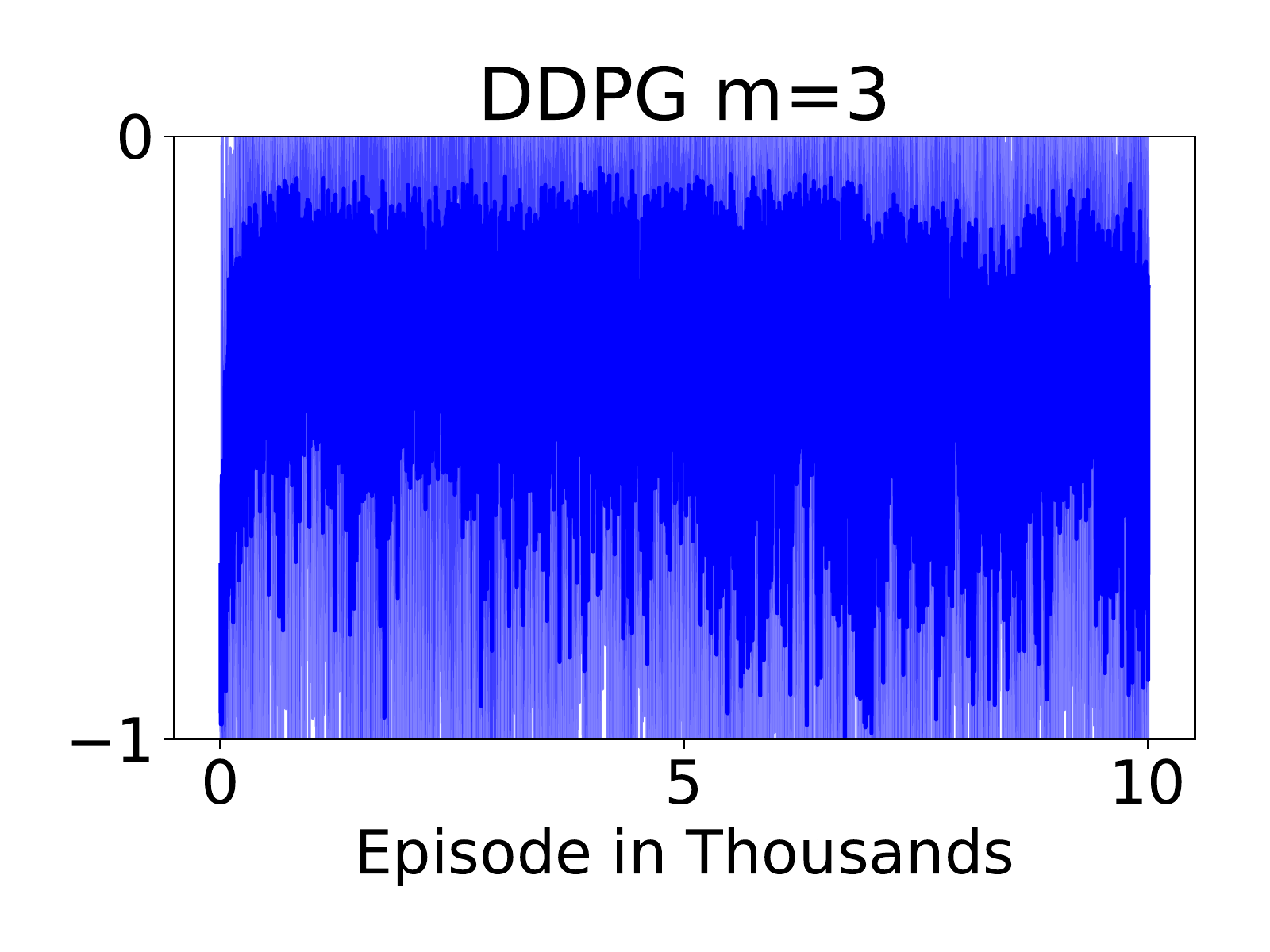}}
	\end{subfigure}
	\caption{Average normalized rewards received during training of 10,000 
	episodes (10 million steps) for each RL algorithm and memory $m$ sizes 1,2 
	and 3. Plots share common $y$ and $x$ axis. Light blue represents \ci confidence 
	interval.  }
	\label{fig:training}
\end{figure*}



\renewcommand\ppoaa{99.9$\pm$0.1}
\renewcommand\ppoab{84.8$\pm$1.6}
\renewcommand\ppoac{100.0$\pm$0.0}
\renewcommand\ppoad{0.0$\pm$0.0}
\renewcommand\ppoae{3.2$\pm$0.4}
\renewcommand\ppoaf{0.0$\pm$0.0}
\renewcommand\ppoaa{65.9$\pm$2.4}
\renewcommand\ppoab{94.1$\pm$4.3}
\renewcommand\ppoac{73.4$\pm$2.7}
\renewcommand\ppoad{113.8$\pm$2.2}
\renewcommand\ppoae{107.7$\pm$2.2}
\renewcommand\ppoaf{128.1$\pm$4.3}
\renewcommand\ppoag{309.9$\pm$7.9}
\renewcommand\ppoah{440.6$\pm$13.4}
\renewcommand\ppoai{215.7$\pm$6.7}
\renewcommand\ppoaj{0.0$\pm$0.0}
\renewcommand\ppoak{0.0$\pm$0.0}
\renewcommand\ppoal{0.0$\pm$0.0}

\renewcommand\ppoba{100.0$\pm$0.1}
\renewcommand\ppobb{53.3$\pm$2.2}
\renewcommand\ppobc{99.2$\pm$0.4}
\renewcommand\ppobd{0.0$\pm$0.1}
\renewcommand\ppobe{11.1$\pm$0.7}
\renewcommand\ppobf{0.2$\pm$0.1}
\renewcommand\ppoba{58.6$\pm$2.5}
\renewcommand\ppobb{125.4$\pm$6.0}
\renewcommand\ppobc{105.0$\pm$5.0}
\renewcommand\ppobd{116.9$\pm$2.5}
\renewcommand\ppobe{103.0$\pm$2.7}
\renewcommand\ppobf{126.8$\pm$3.7}
\renewcommand\ppobg{305.2$\pm$7.9}
\renewcommand\ppobh{674.5$\pm$19.1}
\renewcommand\ppobi{261.3$\pm$7.6}
\renewcommand\ppobj{0.0$\pm$0.0}
\renewcommand\ppobk{0.0$\pm$0.0}
\renewcommand\ppobl{0.0$\pm$0.0}

\renewcommand\ppoca{88.6$\pm$1.4}
\renewcommand\ppocb{47.5$\pm$2.2}
\renewcommand\ppocc{99.3$\pm$0.4}
\renewcommand\ppocd{2.9$\pm$0.5}
\renewcommand\ppoce{45.6$\pm$3.8}
\renewcommand\ppocf{0.3$\pm$0.2}
\renewcommand\ppoca{101.5$\pm$5.0}
\renewcommand\ppocb{128.8$\pm$5.8}
\renewcommand\ppocc{79.2$\pm$3.3}
\renewcommand\ppocd{108.9$\pm$2.2}
\renewcommand\ppoce{94.2$\pm$5.3}
\renewcommand\ppocf{119.8$\pm$2.7}
\renewcommand\ppocg{405.9$\pm$10.9}
\renewcommand\ppoch{1403.8$\pm$58.4}
\renewcommand\ppoci{274.4$\pm$5.3}
\renewcommand\ppocj{0.0$\pm$0.0}
\renewcommand\ppock{0.0$\pm$0.0}
\renewcommand\ppocl{0.0$\pm$0.0}

\renewcommand\trpoaa{42.2$\pm$2.2}
\renewcommand\trpoab{54.1$\pm$2.2}
\renewcommand\trpoac{66.5$\pm$2.1}
\renewcommand\trpoad{48.5$\pm$5.5}
\renewcommand\trpoae{21.8$\pm$2.5}
\renewcommand\trpoaf{29.5$\pm$2.3}
\renewcommand\trpoaa{103.9$\pm$6.2}
\renewcommand\trpoab{150.2$\pm$6.7}
\renewcommand\trpoac{109.7$\pm$8.0}
\renewcommand\trpoad{125.1$\pm$9.3}
\renewcommand\trpoae{110.4$\pm$3.9}
\renewcommand\trpoaf{139.6$\pm$6.8}
\renewcommand\trpoag{1644.5$\pm$52.1}
\renewcommand\trpoah{929.0$\pm$25.6}
\renewcommand\trpoai{1374.3$\pm$51.5}
\renewcommand\trpoaj{-0.4$\pm$0.1}
\renewcommand\trpoak{-0.2$\pm$0.0}
\renewcommand\trpoal{-0.1$\pm$0.0}

\renewcommand\trpoba{43.4$\pm$2.2}
\renewcommand\trpobb{46.4$\pm$2.2}
\renewcommand\trpobc{39.8$\pm$2.1}
\renewcommand\trpobd{44.2$\pm$3.0}
\renewcommand\trpobe{109.3$\pm$12.6}
\renewcommand\trpobf{55.9$\pm$4.4}
\renewcommand\trpoba{161.3$\pm$6.9}
\renewcommand\trpobb{162.7$\pm$7.0}
\renewcommand\trpobc{108.4$\pm$9.6}
\renewcommand\trpobd{100.1$\pm$5.1}
\renewcommand\trpobe{144.2$\pm$13.8}
\renewcommand\trpobf{101.7$\pm$5.4}
\renewcommand\trpobg{1432.9$\pm$47.5}
\renewcommand\trpobh{2375.6$\pm$84.0}
\renewcommand\trpobi{1475.6$\pm$46.4}
\renewcommand\trpobj{0.1$\pm$0.0}
\renewcommand\trpobk{0.4$\pm$0.0}
\renewcommand\trpobl{-0.1$\pm$0.0}

\renewcommand\trpoca{68.0$\pm$2.0}
\renewcommand\trpocb{59.4$\pm$2.2}
\renewcommand\trpocc{82.6$\pm$1.7}
\renewcommand\trpocd{19.1$\pm$1.5}
\renewcommand\trpoce{30.4$\pm$5.0}
\renewcommand\trpocf{12.8$\pm$1.8}
\renewcommand\trpoca{130.4$\pm$7.1}
\renewcommand\trpocb{150.8$\pm$7.8}
\renewcommand\trpocc{129.1$\pm$8.9}
\renewcommand\trpocd{141.3$\pm$7.2}
\renewcommand\trpoce{141.2$\pm$8.1}
\renewcommand\trpocf{147.1$\pm$6.8}
\renewcommand\trpocg{1120.1$\pm$36.4}
\renewcommand\trpoch{1200.7$\pm$34.3}
\renewcommand\trpoci{824.0$\pm$30.1}
\renewcommand\trpocj{0.1$\pm$0.0}
\renewcommand\trpock{-0.1$\pm$0.1}
\renewcommand\trpocl{-0.1$\pm$0.0}

\renewcommand\ddpgaa{59.0$\pm$2.2}
\renewcommand\ddpgab{50.5$\pm$2.2}
\renewcommand\ddpgac{73.9$\pm$1.9}
\renewcommand\ddpgad{29.1$\pm$2.2}
\renewcommand\ddpgae{41.7$\pm$3.6}
\renewcommand\ddpgaf{22.4$\pm$2.4}
\renewcommand\ddpgaa{68.2$\pm$3.7}
\renewcommand\ddpgab{100.0$\pm$5.4}
\renewcommand\ddpgac{79.0$\pm$5.4}
\renewcommand\ddpgad{133.1$\pm$7.8}
\renewcommand\ddpgae{116.6$\pm$7.9}
\renewcommand\ddpgaf{146.4$\pm$7.5}
\renewcommand\ddpgag{1201.4$\pm$42.4}
\renewcommand\ddpgah{1397.0$\pm$62.4}
\renewcommand\ddpgai{992.9$\pm$45.1}
\renewcommand\ddpgaj{0.0$\pm$0.0}
\renewcommand\ddpgak{-0.1$\pm$0.0}
\renewcommand\ddpgal{0.1$\pm$0.0}

\renewcommand\ddpgba{49.2$\pm$1.5}
\renewcommand\ddpgbb{99.1$\pm$4.9}
\renewcommand\ddpgbc{40.7$\pm$1.8}
\renewcommand\ddpgbd{42.0$\pm$5.5}
\renewcommand\ddpgbe{46.7$\pm$8.0}
\renewcommand\ddpgbf{71.4$\pm$7.0}
\renewcommand\ddpgbg{2388.0$\pm$63.9}
\renewcommand\ddpgbh{2607.5$\pm$72.2}
\renewcommand\ddpgbi{1953.4$\pm$58.3}
\renewcommand\ddpgbj{-0.1$\pm$0.0}
\renewcommand\ddpgbk{-0.1$\pm$0.0}
\renewcommand\ddpgbl{-0.0$\pm$0.0}

\renewcommand\ddpgca{85.3$\pm$5.9}
\renewcommand\ddpgcb{124.3$\pm$7.2}
\renewcommand\ddpgcc{105.1$\pm$8.6}
\renewcommand\ddpgcd{101.0$\pm$8.2}
\renewcommand\ddpgce{158.6$\pm$21.0}
\renewcommand\ddpgcf{120.5$\pm$7.0}
\renewcommand\ddpgcg{1984.3$\pm$59.3}
\renewcommand\ddpgch{3280.8$\pm$98.7}
\renewcommand\ddpgci{1364.2$\pm$54.9}
\renewcommand\ddpgcj{0.0$\pm$0.1}
\renewcommand\ddpgck{0.2$\pm$0.1}
\renewcommand\ddpgcl{0.0$\pm$0.0}

\begin{table*}[]
	\footnotesize
\centering
\caption{RL performance evaluation averages from 2,000 command inputs per 
	configuration with \ci confidence where P=PPO, T=TRPO, and D=DDPG.  }
\label{table:averages}
{\setlength{\tabcolsep}{0.2em}
\def\arraystretch{1.15}%
\begin{tabular}{lc|c|c|c|c|c|c|c|c|c|c|c|c|}
\cline{3-14}
\multicolumn{1}{c}{}                        &     & \multicolumn{3}{c|}{Rise (ms)} & \multicolumn{3}{c|}{Peak (\%)} & \multicolumn{3}{c|}{Error (rad/s)} & \multicolumn{3}{c|}{Stability} \\ \cline{2-14} 
\multicolumn{1}{c|}{}                       & $m$ & $\phi$   & $\theta$  & $\psi$  & $\phi$   & $\theta$  & $\psi$  & $\phi$    & $\theta$   & $\psi$    & $\phi$   & $\theta$  & $\psi$  \\ \hline
\multicolumn{1}{|l|}{\multirow{3}{*}{P}}  & 1   & \ppoaa   & \ppoab    & \ppoac  
& \ppoad   & \ppoae    & \ppoaf  & \ppoag    & \ppoah     & \ppoai    & \ppoaj   
& \ppoak    & \ppoal  \\ \cline{2-14} \multicolumn{1}{|l|}{}                      
& 2   & \ppoba   & \ppobb    & \ppobc  & \ppobd   & \ppobe    & \ppobf  & \ppobg    
& \ppobh     & \ppobi    & \ppobj   & \ppobk    & \ppobl  \\ \cline{2-14} 
\multicolumn{1}{|l|}{}                      & 3   & \ppoca   & \ppocb    & 
\ppocc  & \ppocd   & \ppoce    & \ppocf  & \ppocg    & \ppoch     & \ppoci    & 
\ppocj   & \ppock    & \ppocl  \\ \hline
\multicolumn{1}{|l|}{\multirow{3}{*}{T}} & 1   & \trpoaa  & \trpoab   & \trpoac 
& \trpoad  & \trpoae   & \trpoaf & \trpoag   & \trpoah    & \trpoai   & \trpoaj  
& \trpoak   & \trpoal \\ \cline{2-14} \multicolumn{1}{|l|}{}                      
& 2   & \trpoba  & \trpobb   & \trpobc & \trpobd  & \trpobe   & \trpobf & 
\trpobg   & \trpobh    & \trpobi   & \trpobj  & \trpobk   & \trpobl \\ 
\cline{2-14} \multicolumn{1}{|l|}{}                      & 3   & \trpoca  & 
\trpocb   & \trpocc & \trpocd  & \trpoce   & \trpocf & \trpocg   & \trpoch    & 
\trpoci   & \trpocj  & \trpock   & \trpocl \\ \hline
\multicolumn{1}{|l|}{\multirow{3}{*}{D}} & 1   & \ddpgaa  & \ddpgab   & \ddpgac 
& \ddpgad  & \ddpgae   & \ddpgaf & \ddpgag   & \ddpgah    & \ddpgai   & \ddpgaj  
& \ddpgak   & \ddpgal \\ \cline{2-14} \multicolumn{1}{|l|}{}                      
& 2   & \ddpgba  & \ddpgbb   & \ddpgbc & \ddpgbd  & \ddpgbe   & \ddpgbf & 
\ddpgbg   & \ddpgbh    & \ddpgbi   & \ddpgbj  & \ddpgbk   & \ddpgbl \\ 
\cline{2-14} \multicolumn{1}{|l|}{}                      & 3   & \ddpgca  & 
\ddpgcb   & \ddpgcc & \ddpgcd  & \ddpgce   & \ddpgcf & \ddpgcg   & \ddpgch    & 
\ddpgci   & \ddpgcj  & \ddpgck   & \ddpgcl \\ \hline
\end{tabular}}
\end{table*}

\renewcommand\ppoaa{99.8$\pm$0.3}
\renewcommand\ppoab{100.0$\pm$0.0}
\renewcommand\ppoac{100.0$\pm$0.0}
\renewcommand\ppoad{0.1$\pm$0.1}
\renewcommand\ppoae{0.0$\pm$0.0}
\renewcommand\ppoaf{0.0$\pm$0.0}

\renewcommand\ppoba{100.0$\pm$0.0}
\renewcommand\ppobb{53.3$\pm$3.1}
\renewcommand\ppobc{99.8$\pm$0.3}
\renewcommand\ppobd{0.0$\pm$0.0}
\renewcommand\ppobe{20.0$\pm$2.4}
\renewcommand\ppobf{0.0$\pm$0.0}

\renewcommand\ppoca{98.7$\pm$0.7}
\renewcommand\ppocb{74.7$\pm$2.7}
\renewcommand\ppocc{99.3$\pm$0.5}
\renewcommand\ppocd{0.4$\pm$0.2}
\renewcommand\ppoce{5.4$\pm$0.7}
\renewcommand\ppocf{0.2$\pm$0.2}

\renewcommand\trpoaa{32.8$\pm$2.9}
\renewcommand\trpoab{59.0$\pm$3.0}
\renewcommand\trpoac{87.4$\pm$2.1}
\renewcommand\trpoad{72.5$\pm$10.6}
\renewcommand\trpoae{17.4$\pm$3.7}
\renewcommand\trpoaf{9.4$\pm$2.6}

\renewcommand\trpoba{19.7$\pm$2.5}
\renewcommand\trpobb{48.2$\pm$3.1}
\renewcommand\trpobc{56.9$\pm$3.1}
\renewcommand\trpobd{76.6$\pm$5.0}
\renewcommand\trpobe{43.0$\pm$6.5}
\renewcommand\trpobf{38.6$\pm$7.0}

\renewcommand\trpoca{96.8$\pm$1.1}
\renewcommand\trpocb{60.8$\pm$3.0}
\renewcommand\trpocc{73.2$\pm$2.7}
\renewcommand\trpocd{1.5$\pm$0.8}
\renewcommand\trpoce{20.6$\pm$4.1}
\renewcommand\trpocf{20.6$\pm$3.4}

\renewcommand\ddpgaa{84.1$\pm$2.3}
\renewcommand\ddpgab{52.5$\pm$3.1}
\renewcommand\ddpgac{90.4$\pm$1.8}
\renewcommand\ddpgad{11.1$\pm$2.2}
\renewcommand\ddpgae{41.1$\pm$5.5}
\renewcommand\ddpgaf{4.6$\pm$1.0}

\renewcommand\ddpgba{26.6$\pm$2.7}
\renewcommand\ddpgbb{26.1$\pm$2.7}
\renewcommand\ddpgbc{50.2$\pm$3.1}
\renewcommand\ddpgbd{82.7$\pm$8.5}
\renewcommand\ddpgbe{112.2$\pm$12.9}
\renewcommand\ddpgbf{59.7$\pm$7.5}

\renewcommand\ddpgca{39.2$\pm$3.0}
\renewcommand\ddpgcb{44.8$\pm$3.1}
\renewcommand\ddpgcc{60.7$\pm$3.0}
\renewcommand\ddpgcd{52.0$\pm$6.4}
\renewcommand\ddpgce{101.8$\pm$13.0}
\renewcommand\ddpgcf{33.9$\pm$3.4}

\renewcommand\pida{100.0$\pm$0.0}
\renewcommand\pidb{100.0$\pm$0.0}
\renewcommand\pidc{100.0$\pm$0.0}
\renewcommand\pidd{0.0$\pm$0.0}
\renewcommand\pide{0.0$\pm$0.0}
\renewcommand\pidf{0.0$\pm$0.0}

\begin{table}[]
	\footnotesize
\centering
\caption{Success and Failure results for considered algorithms. P=PPO, T=TRPO, and D=DDPG. The row 
	highlighted in blue refers to our best-performing learning agent PPO, while the rows 
        with gray highlight correspond to the best agents for the other two algorithms. }
\label{table:fails}
{\setlength{\tabcolsep}{0.2em}

\def\arraystretch{1.15}%
\begin{tabular}{lc|c|c|c|c|c|c|}
\cline{3-8}
\multicolumn{1}{c}{}                        &     & \multicolumn{3}{c|}{Success 
(\%)} & \multicolumn{3}{c|}{Failure (\%)} \\ \cline{2-8} \multicolumn{1}{c|}{}                       
& $m$ & $\phi$    & $\theta$   & $\psi$   & $\phi$     & $\theta$    & $\psi$    
\\ \hline \rowcolor{blue!30}
\multicolumn{1}{|l|}{\cellcolor{white}\multirow{3}{*}{P}}  & 1   & \ppoaa    & 
\ppoab     & \ppoac   & \ppoad     & \ppoae      & \ppoaf    \\ \cline{2-8} 
\multicolumn{1}{|l|}{}                      & 2   & \ppoba    & \ppobb     & 
\ppobc   & \ppobd     & \ppobe      & \ppobf    \\ \cline{2-8} 
\multicolumn{1}{|l|}{}                      & 3   & \ppoca    & \ppocb     & 
\ppocc   & \ppocd     & \ppoce      & \ppocf    \\ \hline
\multicolumn{1}{|l|}{\cellcolor{white}\multirow{3}{*}{T}} & 1   & \trpoaa   & 
\trpoab    & \trpoac  & \trpoad    & \trpoae     & \trpoaf   \\ \cline{2-8} 
\multicolumn{1}{|l|}{}                      & 2   & \trpoba   & \trpobb    & 
\trpobc  & \trpobd    & \trpobe     & \trpobf   \\ \cline{2-8} 
\rowcolor{gray!30}
\multicolumn{1}{|l|}{\cellcolor{white}}                      & 3   & \trpoca   & 
\trpocb    & \trpocc  & \trpocd    & \trpoce     & \trpocf   \\ \hline
\rowcolor{gray!30}
\multicolumn{1}{|l|}{\cellcolor{white}\multirow{3}{*}{D}} & 1   & \ddpgaa   & 
\ddpgab    & \ddpgac  & \ddpgad    & \ddpgae     & \ddpgaf   \\ \cline{2-8} 
\multicolumn{1}{|l|}{}                      & 2   & \ddpgba   & \ddpgbb    & 
\ddpgbc  & \ddpgbd    & \ddpgbe     & \ddpgbf   \\ \cline{2-8} 
\multicolumn{1}{|l|}{}                      & 3   & \ddpgca   & \ddpgcb    & 
\ddpgcc  & \ddpgcd    & \ddpgce     & \ddpgcf   \\ \hline
\multicolumn{2}{|l|}{PID}                         & \pida     & \pidb      & \pidc    & \pidd      & \pide       & \pidf     \\ \hline
\end{tabular}}
\end{table}

\renewcommand\ppoaa{66.6$\pm$3.2}
\renewcommand\ppoab{70.8$\pm$3.6}
\renewcommand\ppoac{72.9$\pm$3.7}
\renewcommand\ppoad{112.6$\pm$3.0}
\renewcommand\ppoae{109.4$\pm$2.4}
\renewcommand\ppoaf{127.0$\pm$6.2}
\renewcommand\ppoag{317.0$\pm$11.0}
\renewcommand\ppoah{326.3$\pm$13.2}
\renewcommand\ppoai{217.5$\pm$9.1}
\renewcommand\ppoaj{0.0$\pm$0.0}
\renewcommand\ppoak{0.0$\pm$0.0}
\renewcommand\ppoal{0.0$\pm$0.0}

\renewcommand\ppoba{64.4$\pm$3.6}
\renewcommand\ppobb{102.8$\pm$6.7}
\renewcommand\ppobc{148.2$\pm$7.9}
\renewcommand\ppobd{118.4$\pm$4.3}
\renewcommand\ppobe{104.2$\pm$4.7}
\renewcommand\ppobf{124.2$\pm$3.4}
\renewcommand\ppobg{329.4$\pm$12.3}
\renewcommand\ppobh{815.3$\pm$31.4}
\renewcommand\ppobi{320.6$\pm$11.5}
\renewcommand\ppobj{0.0$\pm$0.0}
\renewcommand\ppobk{0.0$\pm$0.0}
\renewcommand\ppobl{0.0$\pm$0.0}

\renewcommand\ppoca{97.9$\pm$5.5}
\renewcommand\ppocb{121.9$\pm$7.2}
\renewcommand\ppocc{79.5$\pm$3.7}
\renewcommand\ppocd{111.4$\pm$3.4}
\renewcommand\ppoce{111.1$\pm$4.2}
\renewcommand\ppocf{120.8$\pm$4.2}
\renewcommand\ppocg{396.7$\pm$14.7}
\renewcommand\ppoch{540.6$\pm$22.6}
\renewcommand\ppoci{237.1$\pm$8.0}
\renewcommand\ppocj{0.0$\pm$0.0}
\renewcommand\ppock{0.0$\pm$0.0}
\renewcommand\ppocl{0.0$\pm$0.0}

\renewcommand\trpoaa{119.9$\pm$8.8}
\renewcommand\trpoab{149.0$\pm$10.6}
\renewcommand\trpoac{103.9$\pm$9.8}
\renewcommand\trpoad{103.0$\pm$11.0}
\renewcommand\trpoae{117.4$\pm$5.8}
\renewcommand\trpoaf{142.8$\pm$6.5}
\renewcommand\trpoag{1965.2$\pm$90.5}
\renewcommand\trpoah{930.5$\pm$38.4}
\renewcommand\trpoai{713.7$\pm$34.4}
\renewcommand\trpoaj{0.7$\pm$0.1}
\renewcommand\trpoak{0.3$\pm$0.0}
\renewcommand\trpoal{0.0$\pm$0.0}

\renewcommand\trpoba{108.0$\pm$8.3}
\renewcommand\trpobb{157.1$\pm$9.9}
\renewcommand\trpobc{47.3$\pm$6.5}
\renewcommand\trpobd{69.4$\pm$7.4}
\renewcommand\trpobe{117.7$\pm$9.2}
\renewcommand\trpobf{126.5$\pm$7.2}
\renewcommand\trpobg{2020.2$\pm$71.9}
\renewcommand\trpobh{1316.2$\pm$49.0}
\renewcommand\trpobi{964.0$\pm$31.2}
\renewcommand\trpobj{0.1$\pm$0.1}
\renewcommand\trpobk{0.5$\pm$0.1}
\renewcommand\trpobl{0.0$\pm$0.0}

\renewcommand\trpoca{115.2$\pm$9.5}
\renewcommand\trpocb{156.6$\pm$12.7}
\renewcommand\trpocc{176.1$\pm$15.5}
\renewcommand\trpocd{153.5$\pm$8.1}
\renewcommand\trpoce{123.3$\pm$6.9}
\renewcommand\trpocf{148.8$\pm$11.2}
\renewcommand\trpocg{643.5$\pm$20.5}
\renewcommand\trpoch{895.0$\pm$42.8}
\renewcommand\trpoci{1108.9$\pm$44.5}
\renewcommand\trpocj{0.1$\pm$0.0}
\renewcommand\trpock{0.0$\pm$0.0}
\renewcommand\trpocl{0.0$\pm$0.0}

\renewcommand\ddpgaa{64.7$\pm$5.2}
\renewcommand\ddpgab{118.9$\pm$8.5}
\renewcommand\ddpgac{51.0$\pm$4.8}
\renewcommand\ddpgad{165.6$\pm$11.6}
\renewcommand\ddpgae{135.4$\pm$12.8}
\renewcommand\ddpgaf{150.8$\pm$6.2}
\renewcommand\ddpgag{929.1$\pm$39.9}
\renewcommand\ddpgah{1490.3$\pm$83.0}
\renewcommand\ddpgai{485.3$\pm$25.4}
\renewcommand\ddpgaj{0.1$\pm$0.1}
\renewcommand\ddpgak{-0.2$\pm$0.1}
\renewcommand\ddpgal{0.1$\pm$0.0}

\renewcommand\ddpgba{nan$\pm$0.0}
\renewcommand\ddpgbb{nan$\pm$0.0}
\renewcommand\ddpgbc{nan$\pm$0.0}
\renewcommand\ddpgbd{0.0$\pm$0.0}
\renewcommand\ddpgbe{0.0$\pm$0.0}
\renewcommand\ddpgbf{0.0$\pm$0.0}
\renewcommand\ddpgbg{2701.9$\pm$90.1}
\renewcommand\ddpgbh{2716.2$\pm$93.2}
\renewcommand\ddpgbi{2569.9$\pm$90.4}
\renewcommand\ddpgbj{-0.0$\pm$0.0}
\renewcommand\ddpgbk{0.0$\pm$0.0}
\renewcommand\ddpgbl{-0.0$\pm$0.0}

\renewcommand\ddpgba{49.2$\pm$2.1}
\renewcommand\ddpgbb{99.1$\pm$6.9}
\renewcommand\ddpgbc{40.7$\pm$2.5}
\renewcommand\ddpgbd{84.0$\pm$10.4}
\renewcommand\ddpgbe{93.5$\pm$15.4}
\renewcommand\ddpgbf{142.7$\pm$12.5}
\renewcommand\ddpgbg{2074.1$\pm$86.4}
\renewcommand\ddpgbh{2498.8$\pm$109.8}
\renewcommand\ddpgbi{1336.9$\pm$50.1}
\renewcommand\ddpgbj{-0.1$\pm$0.0}
\renewcommand\ddpgbk{-0.2$\pm$0.1}
\renewcommand\ddpgbl{-0.0$\pm$0.0}

\renewcommand\ddpgca{73.7$\pm$8.4}
\renewcommand\ddpgcb{172.9$\pm$12.0}
\renewcommand\ddpgcc{141.5$\pm$14.5}
\renewcommand\ddpgcd{103.7$\pm$11.5}
\renewcommand\ddpgce{126.5$\pm$17.8}
\renewcommand\ddpgcf{119.6$\pm$8.2}
\renewcommand\ddpgcg{1585.4$\pm$81.4}
\renewcommand\ddpgch{2401.3$\pm$109.8}
\renewcommand\ddpgci{1199.0$\pm$74.0}
\renewcommand\ddpgcj{-0.1$\pm$0.1}
\renewcommand\ddpgck{-0.2$\pm$0.1}
\renewcommand\ddpgcl{0.1$\pm$0.0}

\renewcommand\pida{79.0$\pm$3.5}
\renewcommand\pidb{99.8$\pm$5.0}
\renewcommand\pidc{67.7$\pm$2.3}
\renewcommand\pidd{136.9$\pm$4.8}
\renewcommand\pide{112.7$\pm$1.6}
\renewcommand\pidf{135.1$\pm$3.3}
\renewcommand\pidg{416.1$\pm$20.4}
\renewcommand\pidh{269.6$\pm$11.9}
\renewcommand\pidi{245.1$\pm$11.5}
\renewcommand\pidj{0.0$\pm$0.0}
\renewcommand\pidk{0.0$\pm$0.0}
\renewcommand\pidl{0.0$\pm$0.0}

\begin{table*}[]
	\footnotesize
\centering
\caption{RL performance evaluation compared to PID of best-performing agent.  
	Values reported are the average of 1,000 command inputs with \ci confidence 
	where P=PPO, T=TRPO, and D=DDPG.  PPO $m=1$ highlighted in blue outperforms 
	all other agents, including PID control.  Metrics highlighted in red for PID 
	control are outpreformed by the PPO agent.}
\label{table:best}
{\setlength{\tabcolsep}{0.2em}
\def\arraystretch{1.15}%
\begin{tabular}{lc|c|c|c|c|c|c|c|c|c|c|c|c|}
\cline{3-14}
 \multicolumn{1}{c}{}                        &     & \multicolumn{3}{c|}{Rise 
(ms)} & \multicolumn{3}{c|}{Peak (\%)} & \multicolumn{3}{c|}{Error (rad/s)} & 
\multicolumn{3}{c|}{Stability} \\ \cline{2-14} \multicolumn{1}{c|}{}                       
& $m$ & $\phi$   & $\theta$  & $\psi$  & $\phi$   & $\theta$  & $\psi$  & $\phi$    
& $\theta$   & $\psi$    & $\phi$   & $\theta$  & $\psi$  \\ \hline 
\rowcolor{blue!30}
\multicolumn{1}{|l|}{\cellcolor{white}\multirow{3}{*}{P}}  & 1   & \ppoaa   & 
\ppoab    & \ppoac  & \ppoad   & \ppoae    & \ppoaf  & \ppoag    & \ppoah     & 
\ppoai    & \ppoaj   & \ppoak    & \ppoal  \\ \cline{2-14} 
\multicolumn{1}{|l|}{}                      & 2   & \ppoba   & \ppobb    & 
\ppobc  & \ppobd   & \ppobe    & \ppobf  & \ppobg    & \ppobh     & \ppobi    & 
\ppobj   & \ppobk    & \ppobl  \\ \cline{2-14} \multicolumn{1}{|l|}{}                      
& 3   & \ppoca   & \ppocb    & \ppocc  & \ppocd   & \ppoce    & \ppocf  & \ppocg    
& \ppoch     & \ppoci    & \ppocj   & \ppock    & \ppocl  \\ \hline
\multicolumn{1}{|l|}{\multirow{3}{*}{T}} & 1   & \trpoaa  & \trpoab   & \trpoac 
& \trpoad  & \trpoae   & \trpoaf & \trpoag   & \trpoah    & \trpoai   & \trpoaj  
& \trpoak   & \trpoal \\ \cline{2-14} \multicolumn{1}{|l|}{}                      
& 2   & \trpoba  & \trpobb   & \trpobc & \trpobd  & \trpobe   & \trpobf & 
\trpobg   & \trpobh    & \trpobi   & \trpobj  & \trpobk   & \trpobl \\ 

\cline{2-14} \rowcolor{gray!30}\multicolumn{1}{|l|}{\cellcolor{white}}                      
& 3   & \trpoca  & \trpocb   & \trpocc & \trpocd  & \trpoce   & \trpocf & 
\trpocg   & \trpoch    & \trpoci   & \trpocj  & \trpock   & \trpocl \\ \hline
\rowcolor{gray!30}
\multicolumn{1}{|l|}{\cellcolor{white}\multirow{3}{*}{D}} & 1   & \ddpgaa  & 
\ddpgab   & \ddpgac & \ddpgad  & \ddpgae   & \ddpgaf & \ddpgag   & \ddpgah    & 
\ddpgai   & \ddpgaj  & \ddpgak   & \ddpgal \\ \cline{2-14} 
\multicolumn{1}{|l|}{}                      & 2   & \ddpgba  & \ddpgbb   & 
\ddpgbc & \ddpgbd  & \ddpgbe   & \ddpgbf & \ddpgbg   & \ddpgbh    & \ddpgbi   & 
\ddpgbj  & \ddpgbk   & \ddpgbl \\ \cline{2-14} \multicolumn{1}{|l|}{}                      
& 3   & \ddpgca  & \ddpgcb   & \ddpgcc & \ddpgcd  & \ddpgce   & \ddpgcf & 
\ddpgcg   & \ddpgch    & \ddpgci   & \ddpgcj  & \ddpgck   & \ddpgcl \\ \hline
\multicolumn{2}{|l|}{PID}                         & \cellcolor{red!30}\pida    
&\cellcolor{red!30} \pidb     & \pidc   & \cellcolor{red!30}\pidd    
&\cellcolor{red!30} \pide     &\cellcolor{red!30} \pidf   &\cellcolor{red!30} 
\pidg     &\cellcolor{red!30} \pidh      &\cellcolor{red!30} \pidi     & \pidj    
& \pidk     & \pidl   \\ \hline
\end{tabular}}
\end{table*}

\subsection{Results}
Each learning agent was trained with an RL algorithm for a total of 10 million 
simulation steps, equivalent to 10,000 episodes or about 2.7 simulation hours.  
The agents configuration is defined as the RL algorithm used for training and 
its memory size $m$.
Training for DDPG took approximately~\rtddpg, while PPO and TRPO took 
approximately~\rtppo and \rttrpo respectively. The average sum of rewards for 
each episode is  normalized between $[-1,0]$ and displayed in 
Figure~\ref{fig:training}. This computed average is from \trials independently 
trained agents with the same configuration, while the 95\% confidence is shown 
in light blue.  Training results show clearly that PPO converges faster than 
TRPO and DDPG, and that also it accumulates higher rewards.  What is also interesting and 
counter-intuitive is that the larger memory size actually \emph{decreases} 
convergence and stability among all trained algorithms.  
Recall from Section~\ref{sec:bg} that RL algorithms learn a policy to map 
states to action.  A reason for the decrease in convergence could be attributed 
to the state space increasing causing the RL algorithm to take longer to learn 
the mapping to the optimal action.  As part of our future work, we plan to 
investigate using separate memory sizes for the error and rotor velocity to 
decrease the state space.  Reward gains during training of TRPO and DDPG are 
quite inconsistent with large confidence intervals. Learning performance  is 
comparable to that observed in\cite{hwangbo2017control}, where TRPO is able 
to accumulate more reward than DDPG during training for the task of navigation 
control.

Each trained agent was then evaluated on 1,000 never before seen command inputs 
in an episodic task. Since there are \trials agents per configuration, each 
configuration was evaluated over a total of 2,000 episodes. The average performance 
metrics for Rise, Peak, Error and Stability for the response to the 2,000 
command inputs is reported in Table~\ref{table:averages}.
Results show that the agent trained with PPO outperforms TRPO and DDPG in every 
measurement.  In fact, PPO is the only one that is able to achieve stability 
(for every $m$), while all other agents have at least one axis where the Stability 
metrics is non-zero. 

Next the best performing agent for each algorithm and memory size is compared  
to the PID controller. The best agent was selected based on the lowest sum of 
errors of all three axis reported by the Error metric. The Success and Failure 
metrics are compared in Table~\ref{table:fails}.
Results show that agents trained with PPO would be the only ones good enough for 
flight, with a success rate close to perfect, and where the roll failure of 
$0.2\%$ is only off by about $0.1\%$ from the setpoint. However the best 
trained agents for TRPO and DDPG are often significantly far away from the 
desired angular velocity. For example TRPO's best agent, $39.2\%$ of the time does 
not reach the desired pitch target with upwards of a $20\%$ error from the setpoint.

Next we provide our thorough analysis comparing the best agents  
in~Table~\ref{table:best}. We have found that RL agents trained with PPO using 
$m=1$ provide performance and accuracy exceeding that of our PID controller in 
regards to rise time, peak velocities achieved, and total error. What is 
interesting is that usually a fast rise time could cause overshoot however the 
PPO agent has on average a faster rise time and less overshoot. Both PPO and PID 
reach a stable state measured halfway through the simulation.

\begin{figure*}
	\includegraphics[width=\textwidth]{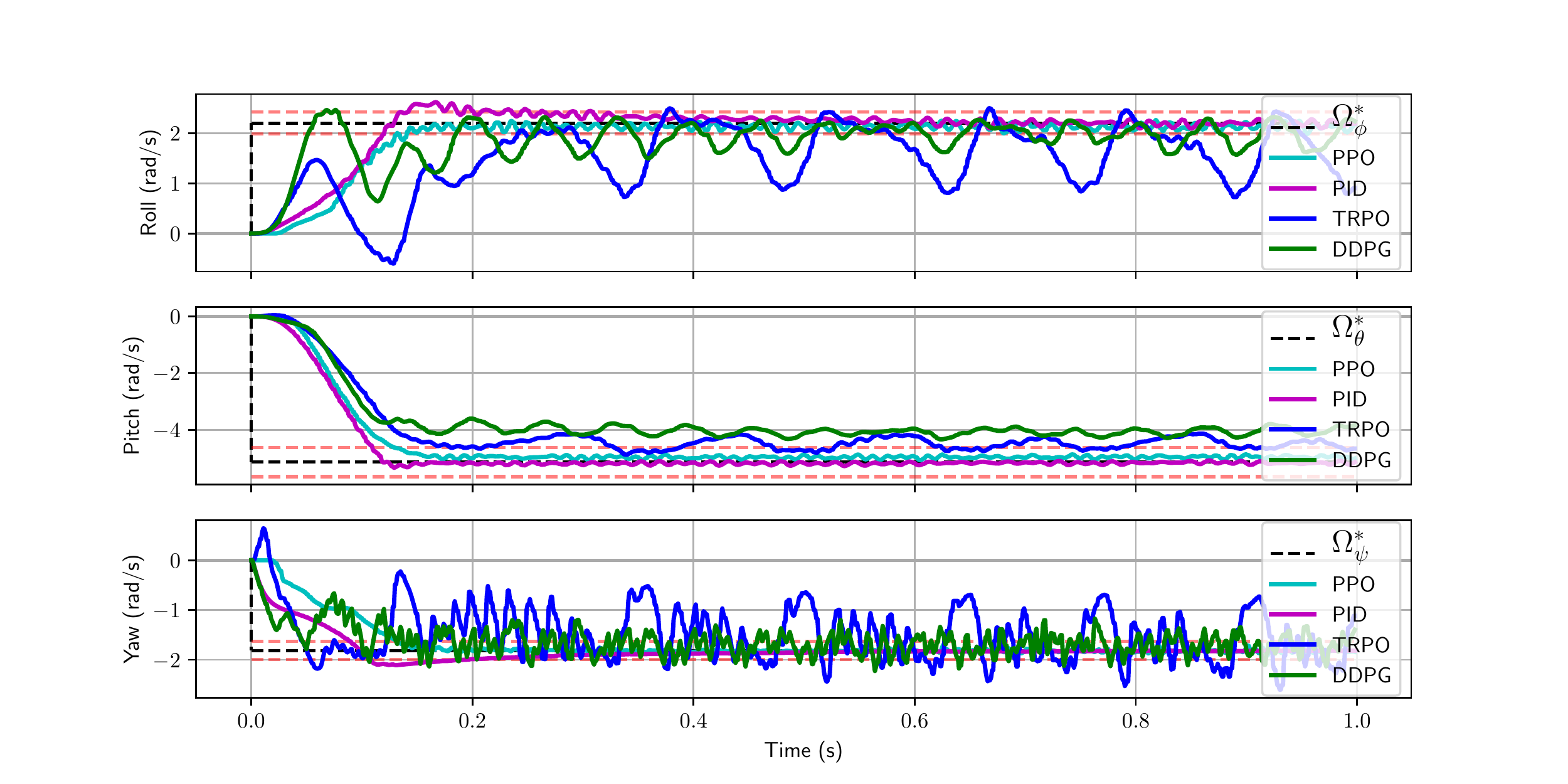}
	\caption{Step response of best trained RL agents compared to
          PID. Target angular velocity is $\Omega^*=[2.20, -5.14,
            -1.81]$ rad/s shown by dashed black line.  Error bars
          $\pm$\thresholdband of initial error from $\Omega^*$ are
          shown in dashed red.}
	\label{fig:stepcompare}
\end{figure*}

\begin{figure*}
	\includegraphics[width=\textwidth]{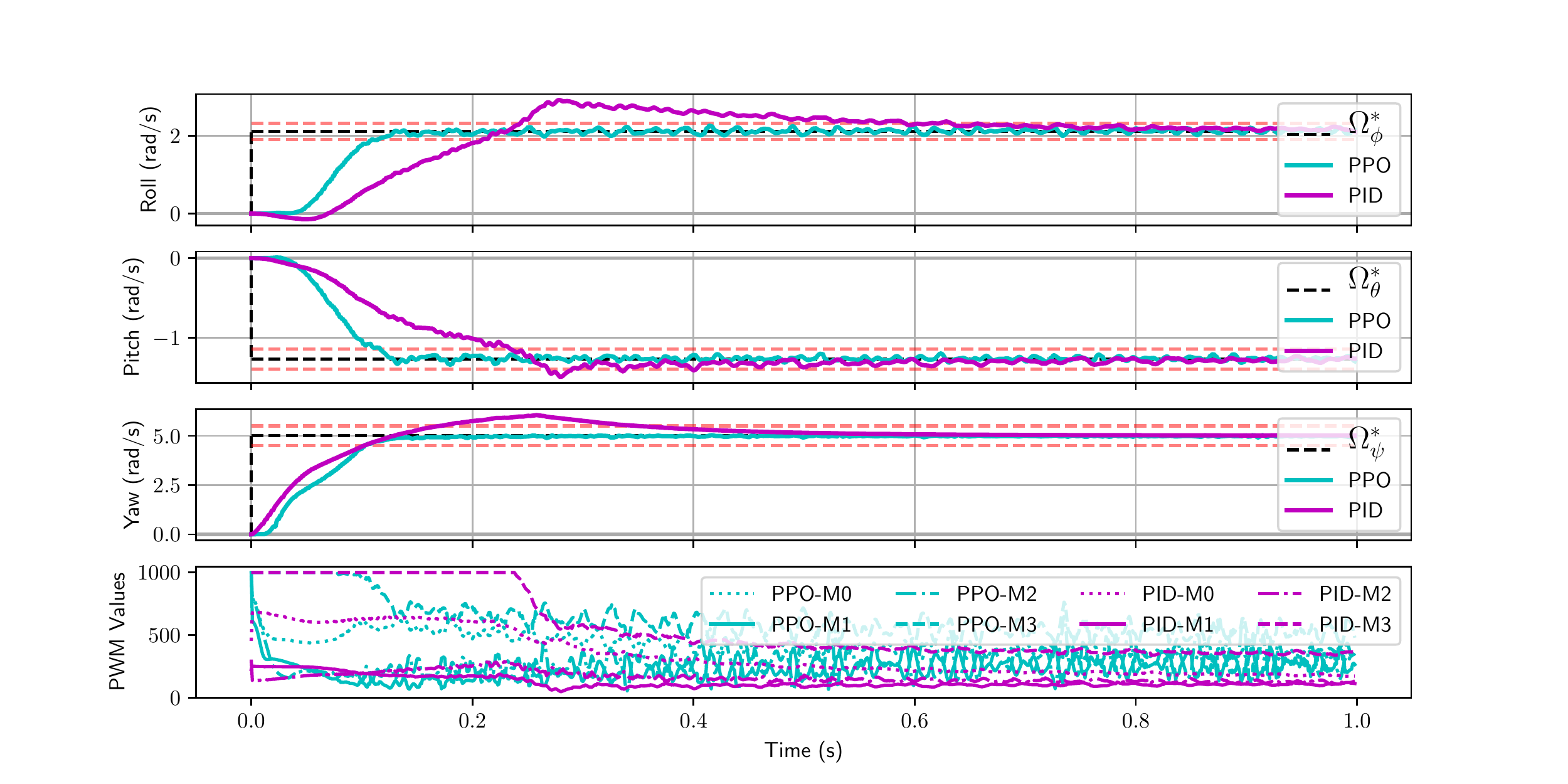}
	\caption{Step response and PWM motor signals of the best trained PPO agent 
		compared to PID.  Target angular velocity is  $\Omega^*=[2.11, -1.26, 
		5.00]$ rad/s shown by dashed black line.  Error bars $\pm$\thresholdband of 
		initial error from $\Omega^*$ are shown in dashed red.}
	\label{fig:bestcompare}
\end{figure*}

To illustrate the performance of each of the best agents a random simulation is 
sampled and the step response for each attitude command is displayed in 
Figure~\ref{fig:stepcompare} along with the target angular velocity to achieve 
$\Omega^*$. All algorithms reach some steady state however only PPO and PID do 
so within the error band indicated by the dashed red lines.  TRPO and DDPG have 
extreme oscillations in both the roll and yaw axis, which would cause 
disturbances during flight.  To highlight the performance and accuracy of the 
PPO agent we sample another simulation and show the step response and also the 
PWM control signals generated by each controller in 
Figure~\ref{fig:bestcompare}. In this figure we can see the PPO agent has 
exceptional tracking capabilities of the desired attitude. The PPO agent has a  
2.25 times faster rise time on the roll axis, 2.5 times faster on the pitch axis 
and 1.15 time faster on the yaw axis.
Furthermore the PID controller experiences slight overshoot in both the roll and 
yaw axis while the PPO agent does not. In regards to the control output, the PID 
controller exerts more power to motor three but then motor values eventually 
level off while the PPO control signal oscillates comparably more.

\section{Future Work and Conclusion}
\label{sec:future}
\label{sec:conclusion}
In this paper we presented our RL training environment \gym for developing 
intelligent attitude controllers for UAVs. We placed an emphasis on digital 
twinning concepts to allow transferability to real hardware. We used \gym to 
evaluate the performance of state-of-the-art RL algorithms PPO, TRPO and DDPG to 
identify if they are appropriate to synthesize high-precision attitude 
flight controllers.  Our results highlight that: (i) 
RL can train accurate attitude controllers; and (ii) that those trained with PPO outperformed 
a fully tuned PID controller on almost every metric. 
Although we base our evaluation on results obtained in episodic tasks, 
we found that trained agents were able to perform exceptionally well also in 
continuous tasks without retraining~\extra.  This suggests that training using 
episodic tasks is sufficient for developing intelligent attitude controllers.  
The results presented in this work can be considered as a first
milestone and a good motivation to further inspect the boundaries of RL 
for intelligent control. With this premise, we plan to develop
our future work along three main avenues. On the one hand, we plan to  
investigate and harness the true power of RL's ability to adapt and learn in environments
with dynamic properties (e.g. wind, variable payload). On the other hand we intend
to transfer our trained agents onto a real aircraft to evaluate their live
performance. Furthermore, we plan to expand \gym to support 
other aircraft such as fixed wing, while continuing to increase the realism of the simulated
environment by improving the accuracy of our digital twins.

\balance
\bibliographystyle{IEEEtran}

\bibliography{references}
\appendices
\section{Continuous Task Evaluation}
\label{sec:continuous}

\begin{figure*}
	\centering
	{\includegraphics[trim=55 0 65 40, clip, width=\textwidth]{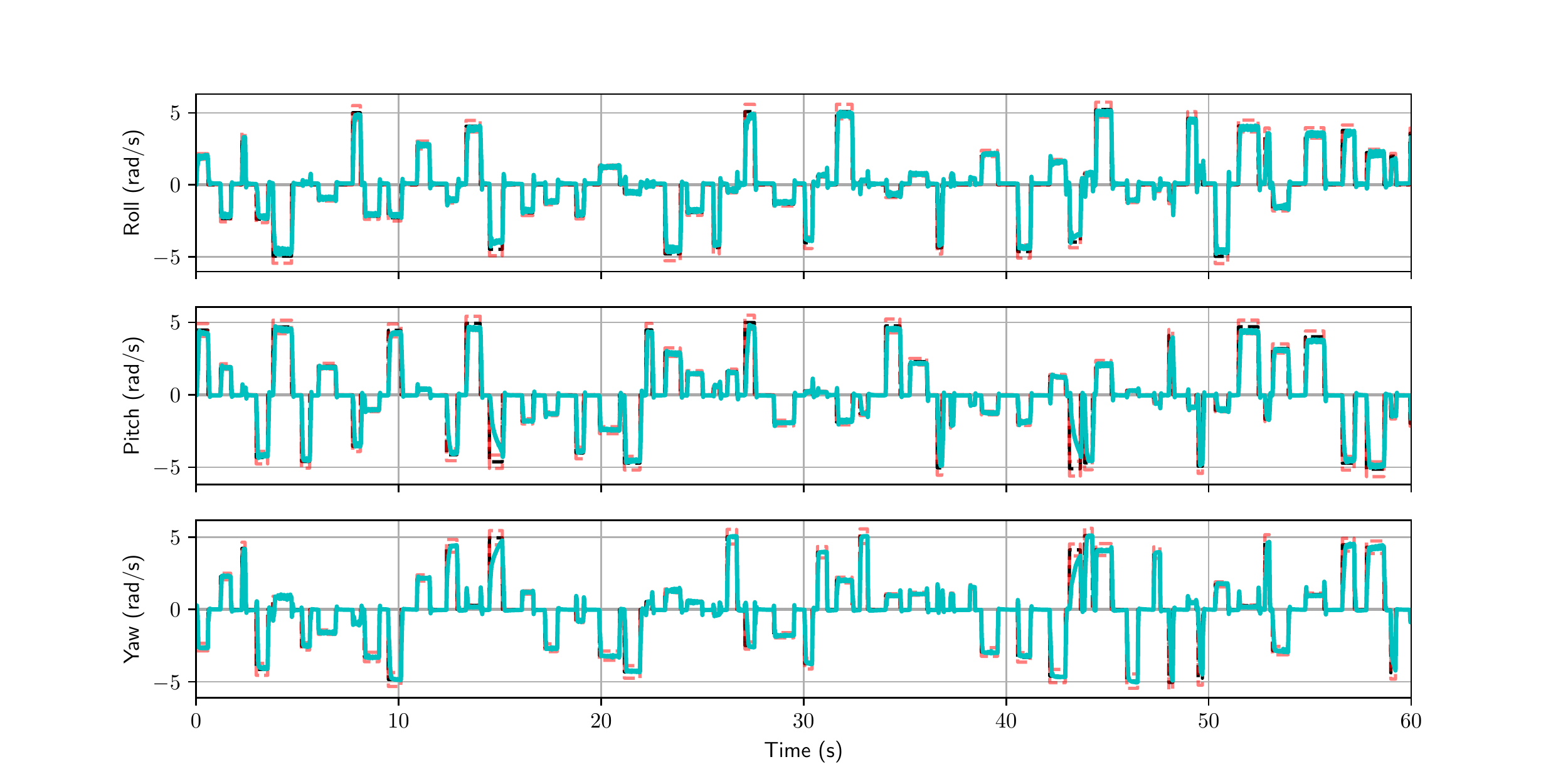}}
	\caption{Performance of PPO agent trained with episodic tasks but evaluated 
	using a continuous task for a duration of 60 seconds. The time in seconds at which a 
	new command is issued is randomly sampled from the interval $[0.1,1]$ and each issued command
	is maintained for a random duration also sampled from $[0.1, 1]$.  Desired 
	angular velocity is specified by the black line while the red line is the 
	attitude tracked by the agent.}
	\label{fig:ppo1}
\end{figure*}

\begin{figure*}
	\centering
	{\includegraphics[trim=45 0 65 30, clip, width=\textwidth]{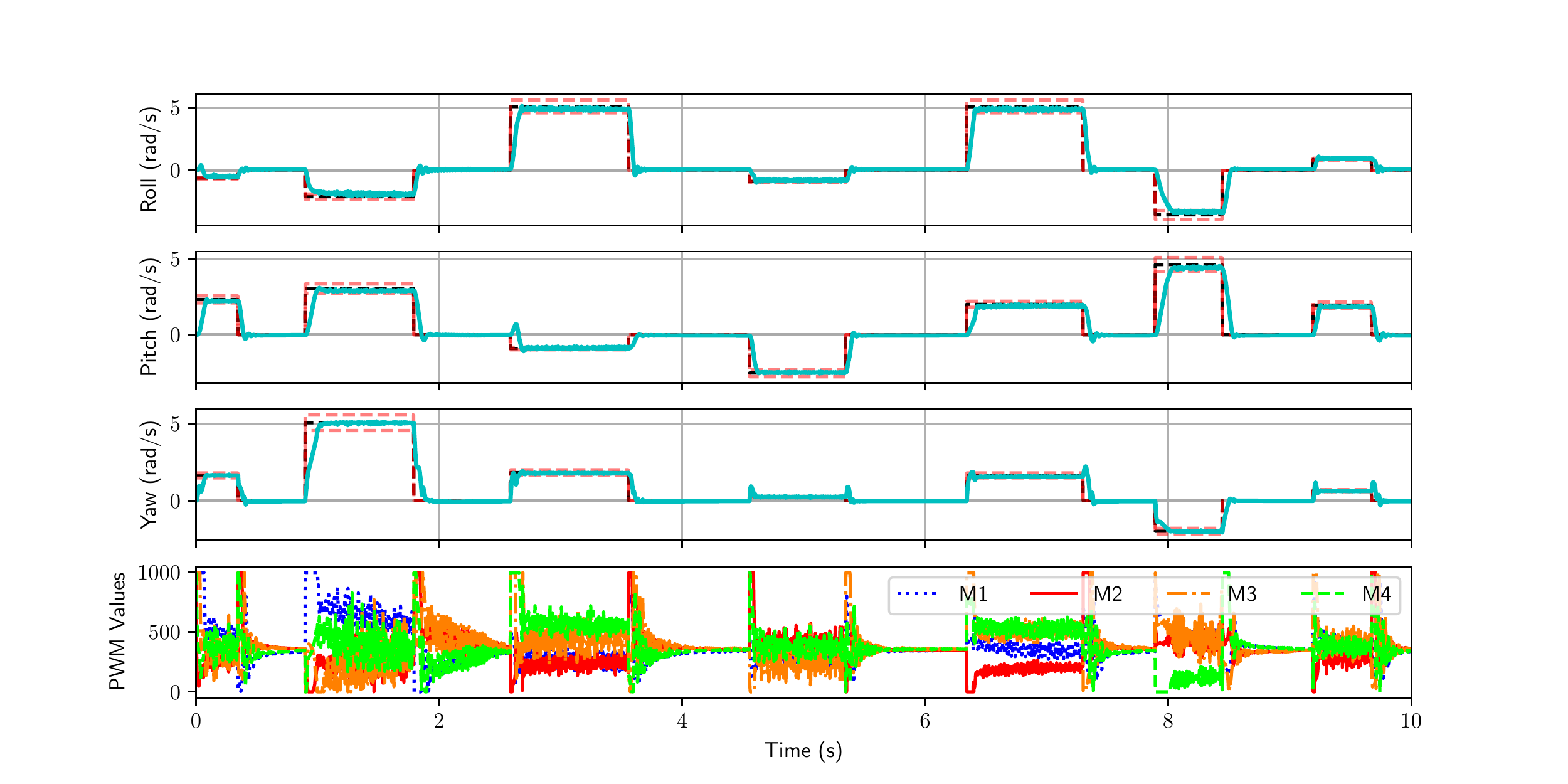}}
	\caption{Close up of continuous task results for PPO agent with PWM values.}
	\label{fig:ppo2}
\end{figure*}

\begin{sidewaysfigure*}
	\centering
	{\includegraphics[trim=45 0 65 30, clip, width=\textwidth]{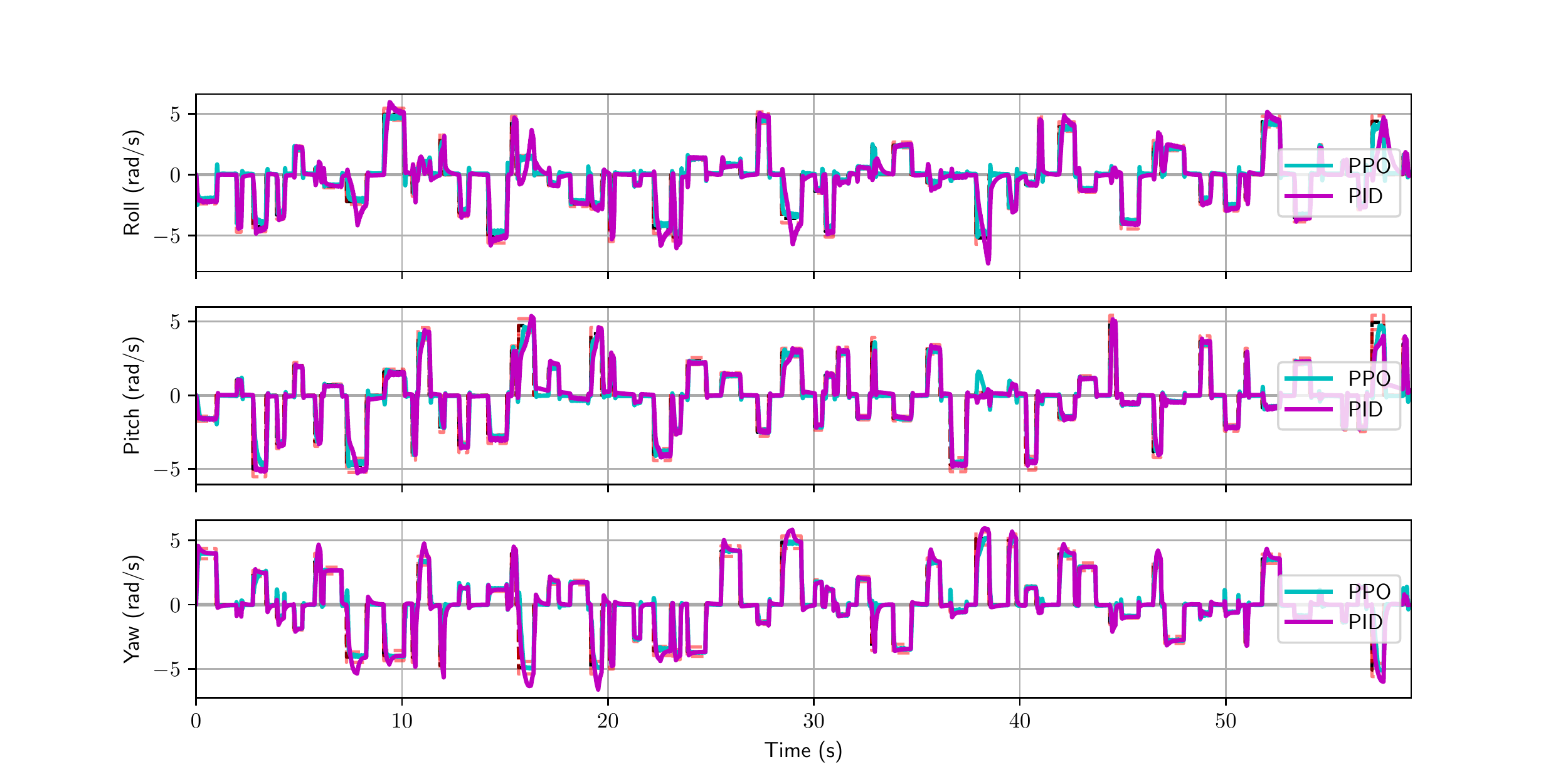}}
	\caption{Response comparison of a PID and PPO agent evaluated in continuous 
		task environment.  The PPO agent, however, is only trained using 
		episodic tasks.}
	\label{fig:ppopid}
\end{sidewaysfigure*}

In this section we briefly expand on our findings that show that even  if agents 
are trained through episodic tasks their performance transfers to continuous 
tasks without the need for additional training.  
Figure~\ref{fig:ppo1}  shows that an agent trained with Proximal Policy 
Optimization~(PPO) using episodic tasks has exceptional performance when 
evaluated in a continuous task. Figure~\ref{fig:ppo2} is a close up of another 
continuous task sample showing the details of the tracking and corresponding 
motor output. These results are quite remarkable as they suggest that training 
with episodic tasks is sufficient for developing intelligent attitude flight 
controller systems capable of operating in a continuous environment. In 
Figure~\ref{fig:ppopid} another continuous task is sampled and the PPO agent is 
compared to a PID agent. The performance evaluation shows the PPO agent to have 
22\% decrease in overall error in comparison to the PID agent.

\end{document}